\documentclass[twocolumn,10pt]{asme2ej}

\usepackage{graphicx} 
\usepackage{amsmath}
\usepackage{amssymb}
\usepackage{hyperref} 
\hypersetup{
	colorlinks=true,
	linkcolor=blue,
	citecolor=blue,
	urlcolor=blue,
}
\usepackage[square,numbers,sort&compress]{natbib} 
\usepackage{bm}      
\usepackage{makecell} 

\usepackage{subcaption} 
\usepackage[labelsep=period]{caption} 

\graphicspath{{fig/}}

\title{Dual-domain fused LSTM modeling for efficient time-dependent reliability analysis}

\author{Yixin Zhang
    \affiliation{   
    School of Aeronautics and Astronautics\\
    Sun Yat-sen University\\
    Shenzhen, Guangdong 518106, China\\
    Email: zhangyx565@mail2.sysu.edu.cn
    }
}

\author{Mingyang Li\thanks{Corresponding author.}
    \affiliation{
    School of Aeronautics and Astronautics\\
    Sun Yat-sen University\\
    Shenzhen, Guangdong 518106, China\\
    Email: limingy3@mail.sysu.edu.cn
    }
}

\author{Zichao Jiang
    \affiliation{
    School of Aeronautics and Astronautics\\
    Sun Yat-sen University\\
    Shenzhen, Guangdong 518106, China\\
    Email: jiangzch8@mail.sysu.edu.cn
    }
}

\begin{document}

\maketitle

\begin{abstract}
Time-dependent reliability analysis is crucial for ensuring the long-term safety and performance of engineering systems under uncertainties. However, traditional surrogate model methods often struggle to incorporate time-independent random variables and capture their complex interactions with time-dependent stochastic processes. To overcome this limitation, this paper proposes a dual-domain fused long short-term memory (DDF-LSTM) model for efficient and accurate time-dependent reliability analysis. A novel network architecture is developed to jointly process information from both time-dependent and time-independent domains. Specifically, the time-independent variables are embedded into the initial hidden states, and a fully connected layer is introduced to map both LSTM outputs and time-independent variables into the final output space. Furthermore, an improved loss function is designed to emphasize the model's sensitivity to minimum responses, thereby improving the precision of failure probability estimation. The proposed method effectively captures the dependencies among random variables, stochastic processes, and the temporal behavior of limit state functions. Once trained, the DDF-LSTM model enables efficient Monte Carlo simulation to estimate time-dependent failure probabilities with minimal computational cost. Four case studies validate the proposed method's enhanced computational efficiency and predictive accuracy.

\end{abstract}

\begin{nomenclature}
\entry{$\mathbf{X}$}{time-independent random variables}
\entry{$\mathbf{Z}(t)$}{time-dependent stochastic processes}
\entry{$P_f$}{probability of failure}
\entry{$z_G(t)$}{Gaussian process}
\entry{$\mu_G(t)$}{mean function}
\entry{$\sigma_G(t)$}{standard deviation function}
\entry{$\rho_G(t)$}{auto correlation function}
\entry{$\mathrm{Cov}$}{covariance}
\entry{$\mathbf{\Sigma}$}{covariance matrix}
\entry{$\mathbf{Q}$}{matrix of eigenvectors}
\entry{$\mathbf{\Lambda}$}{diagonal matrix of eigenvalues}
\entry{$m$}{number of dominated eigenfunctions}
\entry{$\lambda$}{eigenvalues}
\entry{$\bm{p}$}{uncorrelated standard normal random variables}
\entry{$\bm{C}_t$}{cell state}
\entry{$\bm{h}_t$}{hidden state}
\entry{$\bm{c}_0$}{initial cell state}
\entry{$\bm{h}_0$}{initial hidden state}
\entry{$\mathbf{M}$}{time-dependent input matrix}
\entry{$\widehat{\mathbf{Y}}$}{global prediction matrix}
\entry{$\widehat{\mathbf{Y}}^*$}{response predictions of MCS samples}
\entry{$I_f$}{indicator function}
\end{nomenclature}
 
\section{Introduction}

Reliability quantifies the probability of an engineering system successfully executing its design-specified functions under multi-source uncertainties throughout its service life \cite{wang2023time}. Accurate and efficient reliability prediction at the preliminary design stage is essential for lifecycle cost estimation, maintenance strategy formulation, and system performance optimization. Traditional reliability analysis methods typically assume time-independent failure probabilities, as they primarily account for time-independent uncertainties arising from input parameter variations and insufficient data. However, in practical engineering applications, many uncertainties exhibit temporal dependence, including material performance degradation, corrosion, and stochastic dynamic loads. These time-dependent factors generally cause system reliability to degrade over time \cite{jia2021stochastic,alkaff2021discrete}. Therefore, performing time-dependent reliability analysis is critical to ensure long-term structural safety and optimize system performance throughout its service life. Time-dependent reliability analysis primarily focuses on assessing the temporal evolution of system or component failure probabilities. Considering time parameters and stochastic processes increases the complexity of reliability evaluation \cite{wang2024subdomain,dang2025towards}.

Research efforts \cite{zhao2022nested,guo2024probability,li2025novel,yu2023adaptive,zhang2017time,wu2021time} have been devoted to addressing these challenges, which can be roughly categorized into outcrossing rate methods and extreme value methods. The outcrossing rate methods define failure as a crossing event where the system response exceeds the limit state, and the temporal evolution of system reliability is mathematically represented through the out-crossing rate. Under the assumption of independent crossing events, time-dependent failure probabilities are derived from the cumulative integration of this rate. The PHI2 method \cite{andrieu2004phi2} utilizes the first-order reliability method (FORM) to reduce computational burden during outcrossing rate evaluation. Zhang et al. \cite{zhang2021efficient} enhanced efficiency with the moment-based PHI2 method, decoupling finite element analysis from the reliability cycles by pre-defining limit state components as random variables and estimating their statistical properties. Based on PHI2 method, Li et al. \cite{li2022explicit} eliminate the two-dimensional numerical integration by deriving an explicit out-crossing rate model, which consistently reduces computational time especially when dealing with long forecast time interval. Recently, significant efforts have been devoted to developing analytical approaches for time-dependent assessment involving non-stationary and non-Gaussian stochastic processes \cite{zhang2024improved,yang2023time,li2024outcrossing}. Although notable advancements have been made in outcrossing rate methods over the past decades, their implementation in practical time-dependent reliability evaluation continues to face a major constraint: their accuracy cannot be guaranteed when outcrossing events among different time instants are highly dependent. 
The extreme-value approach reformulates the problem into time-independent counterparts, in which the distribution of the extreme response is of critical importance to estimate the time-dependent reliability \cite{zhang2023single,zhang2024loading}. Meng et al. \cite{meng2021efficient} combined stochastic process discretization and FORM for approximating the time-invariant limit state function, then sampling-based method \cite{zha2025single,xin2025new,guo2023adaptive,cao2022single,yun2022coupled,hong2023new} is utilized for estimating the cumulative failure probability. As the most straightforward way to construct the extreme value distribution, various sampling strategies have been applied for extreme value methods. With the combination of the Hermite polynomial model and importance sampling, Zhao et al. \cite{zhao2022time} transformed the limit state function into a moment-based equivalent Gaussian process, then the time-dependent reliability is approximated through solving a multi-dimensional Gaussian integral. Yuan et al. \cite{yuan2024adaptive} developed an adaptive strategy to determine the best setting of importance sampling components, and the computational costs is reduced through forming an overall estimator based on an optimal combination algorithm. The major challenge for extreme value methods lies in that it can be computationally expensive due to the global optimization process and the requirement of a large amount of extreme value samples.

To alleviate the computational burden and improve the precision of reliability estimation, a variety of surrogate modeling techniques have been utilized for solving time-dependent reliability problems, including Support Vector Machine (SVM) \cite{pepper2022adaptive,zhang2025efficient}, Polynomial Chaos Expansion (PCE) \cite{hawchar2017principal}, Artificial Neural Network (ANN) \cite{liu2022artificial,ren2022ensemble}. Among them, Gaussian Process Regression \cite{teixeira2021adaptive,song2022estimation,yang2023constraint} has gained prominence due to its statistical foundation, which naturally supports adaptive learning strategies for iterative model refinement. For time-dependent reliability analysis, adaptive Kriging methods can be broadly categorized by their sampling loop structure. A common strategy is the double-loop approach, where an outer loop manages the surrogate model's global accuracy, while an inner loop addresses the specific challenges of the temporal dimension. For instance, Wang et al. \cite{wang2012nested} established the Nested Extreme Response Surface (NERS) framework, employing Efficient Global Optimization in an inner loop to identify extreme responses, and building a surrogate model in the outer loop. Wang et al. \cite{wang2017confidence} introduced the Adaptive Extreme Response Surface (AERS) method, which constructs a Kriging model to predict limit state values at discrete time points. This method iteratively improves accuracy through confidence-based adaptive sampling, ensuring computational efficiency and high precision. Hu et al. \cite{hu2015mixed} developed a mixed efficient global optimization framework that enhances computational efficiency by concurrently sampling random variables and time, while incorporating an adaptive Kriging approach. These methods can effectively address time-dependent reliability problems but still require substantial computational resources due to the double-loop structure. To mitigate the high computational cost, significant efforts have been devoted to developing single-loop methods, Hu and Mahadevan \cite{hu2016single} introduced a single-loop Kriging modeling method that sequentially refines the surrogate model by simultaneously selecting sample points for both random variables and time parameters. Wang et al. \cite{wang2016time} proposed the equivalent stochastic process transformation (eSPT) method, which converts stochastic processes and time into random variables, thereby reformulating the time-dependent problem into a time-independent one with reduced dimensionality.

More recently, Deep Learning techniques, specifically Recurrent Neural Networks (RNNs) and Long Short-Term Memory (LSTM) networks, have emerged as powerful tools for processing sequence data due to their ability to capture long-term temporal dependencies. While LSTMs are adept at modeling dynamic stochastic processes, a significant challenge remains in effectively integrating time-independent (static) random variables with time-dependent inputs. Existing hybrid frameworks often employ multiple sub-models (e.g., combining LSTM with Feedforward Neural Networks) \cite{li2022lstm}, which increases architectural complexity and potential error accumulation. Other approaches, such as the rLSTM-AE \cite{zhang2024rlstm}, handle static variables by duplicating them across all time steps to create a "pseudo" time-series. However, this concatenation strategy introduces redundant information and feature collinearity, which can hinder the learning of nonlinear couplings between static and dynamic uncertainties, and more critically, lead to unstable gradient propagation during training and biased reliability prediction.

To overcome these limitations, this study proposes a Dual-Domain Fused LSTM (DDF-LSTM) framework for efficient time-dependent reliability analysis. The proposed architecture introduces a novel fusion mechanism where time-independent variables are embedded directly into the initial hidden and cell states of the LSTM. This design provides a consistent contextual foundation for the temporal processing, reflecting the persistent influence of static parameters. Furthermore, a post-processing fully connected layer fuses the LSTM's temporal features with the original static inputs to ensure information integrity. To enhance the estimation of low-probability failure events, a minimum-response-sensitive loss function is designed. This end-to-end architecture enables the unified modeling of static and dynamic uncertainties, facilitating efficient Monte Carlo Simulation (MCS) for time-dependent failure probability estimation with superior computational efficiency and robustness.

This paper is organized as follows. Section~\ref{sect_background} reviews the theoretical foundations of time-variant reliability assessment and stochastic process modeling. Section~\ref{sect_method} details the methodology and implementation of the proposed DDF-LSTM approach. Section~\ref{sect_case} presents four case studies that demonstrate the model's performance in terms of accuracy, efficiency, and robustness. Finally, Section~\ref{sect_conclusion} concludes the study.

\section{Time-dependent reliability analysis}
\label{sect_background}

\subsection{Problem Statement}
In engineered systems, time-dependent uncertainties may be introduced due to load fluctuations, system performance degradation, and aging issues. Therefore, the performance of engineered systems is often characterized as a complex stochastic process. These uncertainties are typically modeled using a combination of random variables $\mathbf{X}$ and time-dependent stochastic processes $\mathbf{Z}(t)$. In general, a limit state function, $G(\mathbf{X},\mathbf{Z}(t),t)$, is commonly used to describe system behavior under uncertainty, where the failure surface $G(\mathbf{X},\mathbf{Z}(t),t)=0$ delineates the boundary between safe and failure domains. The failure domain encompasses all scenarios where the system response fails to meet the required criteria. A failure event is considered to occur if the system response falls below the threshold at any time during its life cycle $[0,T]$, which is expressed as:

\begin{equation}
G(\mathbf{X},\mathbf{Z}(t),t)<0, \quad \exists t \in [0,T]
\end{equation}

Thus, the probability of failure $P_f$ is determined by integrating the joint probability density function over the failure domain, defined as

\begin{equation}
P_f(0,T)=P\{G(\mathbf{X},\mathbf{Z}(t),t)<0, \exists t \in [0,T]\}
\label{eq_Pf}
\end{equation}

Given the intrinsic time-dependent nature of system failure, closed-form solutions to Eqn.~(\ref{eq_Pf}) are generally intractable. Consequently, considerable research efforts have been devoted to addressing this challenge. Among them, sampling-based methods combined with surrogate modeling techniques have emerged as effective solutions, significantly reducing the computational cost associated with direct failure probability evaluation.

\subsection{Spectral Decomposition of Stochastic Processes}
A stochastic process is a collection of random variables indexed by time or space, describing the evolution laws of dynamic uncertain systems. To facilitate numerical analysis, discretization methods based on spectral decomposition have been developed to simulate such processes using a finite set of uncorrelated random variables, such as the Karhunen-Loève expansion and the Expansion Optimal Linear Estimation (EOLE) \cite{sudret2002comparison,zhang1994orthogonal}. Among these, the EOLE technique is employed in this study due to its computational efficiency for generating realizations at discrete time nodes, which aligns perfectly with the data structure required for LSTM training. For example, a Gaussian process $z_G(t)$ is completely specified through three deterministic functions: mean function $\mu_Z(t)$, standard deviation function $\sigma_Z(t)$, and autocorrelation function $\rho_Z(t)$. Upon discretizing the temporal domain into $s$ nodes, the covariance between two time points $t_i$ and $t_j$ is given by:
\begin{equation}
Cov(t_i,t_j) =\sigma_Z(t_i)\sigma_Z(t_j) \rho_Z(t_i,t_j)
\end{equation}

Using this expression, the covariance matrix $\mathbf{\Sigma}$ over all time nodes can be constructed as:
\begin{equation}
\mathbf{\Sigma} = 
\begin{bmatrix}
Cov(t_1,t_1) & Cov(t_1,t_2) & \cdots & Cov(t_1,t_s) \\
Cov(t_2,t_1) & Cov(t_2,t_2) & \cdots & Cov(t_2,t_s) \\
\vdots & \vdots & \ddots & \vdots \\
Cov(t_s,t_1) & Cov(t_s,t_2) & \cdots & Cov(t_s,t_s)
\end{bmatrix}
\end{equation}

Employing spectral decomposition, the covariance matrix $\mathbf{\Sigma}$ decomposes into $\mathbf{Q} \mathbf{\Lambda} \mathbf{Q}^{\!\mathrm{T}}$, where matrix $\mathbf{Q}$ assembles eigenvectors $[\bm{Q}_1, \bm{Q}_2, ..., \bm{Q}_s]$, and $\mathbf{\Lambda}$ is a diagonal matrix of eigenvalues. Consequently, the Gaussian process is approximated as
\begin{equation}
z_G(t) \approx \mu_Z(t)+ \sum_{i=1}^{m} \sqrt{\lambda_i} Q_i(t)p_i
\end{equation}
where $\lambda_i$ are the eigenvalues, $p_i$ are independent standard normal variables, and $m$ is the number of truncated terms. Non-Gaussian process simulation techniques are detailed in references \cite{phoon2005simulation,deodatis2025spectral}.

\section{Dual-domain fused LSTM for time-dependent reliability assessment}
\label{sect_method}

This section introduces a DDF-LSTM approach for the analysis of time-dependent reliability. The fundamental concepts of the LSTM architecture are first introduced in Section~\ref{sect_LSTM}. The proposed DDF-LSTM methodology is then elaborated in Section~\ref{sect_DDF-LSTM}. The complete implementation procedure for reliability approximation is described in Section~\ref{sect_procedure}.

\subsection{Principles of LSTM}
\label{sect_LSTM}
\begin{figure}
\centerline{\includegraphics[width=\columnwidth]{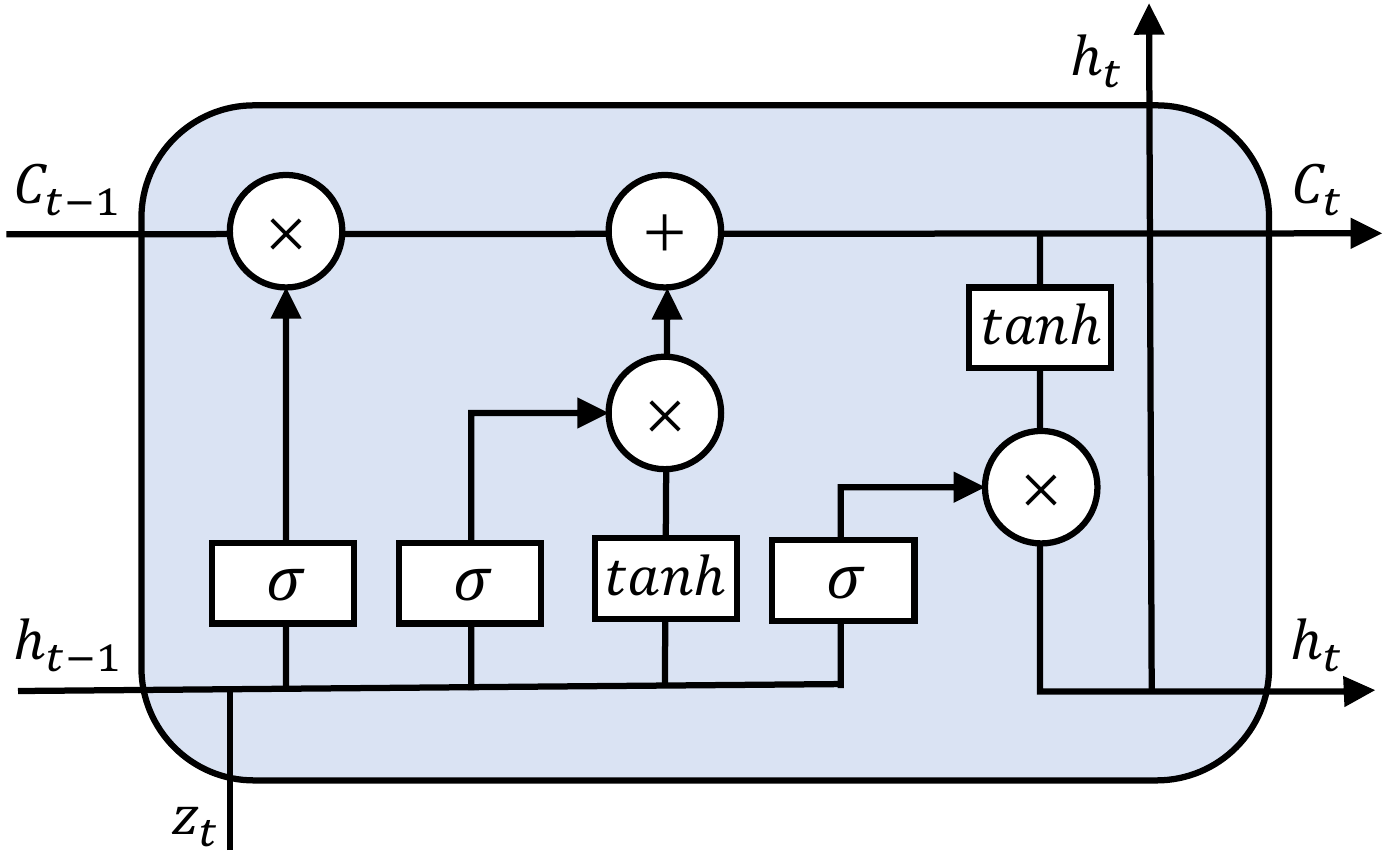}}
\caption{LSTM cell structure diagram}
\label{fig1}
\end{figure}

As a typical network for processing time-series data, RNN faces challenges in learning long-term temporal dependencies due to gradient vanishing and exploding issues during training. LSTM networks were developed to overcome these issues by introducing memory units and gate mechanisms. As illustrated in Fig.~\ref{fig1}, LSTM employs three gates—forget, input, and output—to regulate the information flow, enabling selective retention or discard of information, which is ultimately reflected in the cell state $\bm{C}_t$ and output signal $\bm{h}_t$.

The forget gate regulates which information from the previous cell state $\bm{C}_{t-1}$ should be forgotten. At each time step, the previous hidden state $\bm{h}_{t-1}$ and the current input $\bm{z}_t$ are processed through weight matrices and biases, followed by a sigmoid activation to produce the forget gate output:
\begin{equation}
\bm{f}_t=\sigma(\mathbf{W}_{hf} \bm{h}_{t-1}+\mathbf{W}_{zf} \bm{z}_t+\bm{b}_f)
\end{equation}
where $\bm{f}_t$ denotes the output of the forget gate, $\mathbf{W}_{hf}$, $\mathbf{W}_{zf}$, and $\bm{b}_f$ represent recurrent weights, input weights, and bias, respectively. The sigmoid function $\sigma (.)$ guarantees that the output values remain constrained within the range of 0 to 1, effectively controlling the memory update. 
Next, the input gate identifies which new information should be added to the cell state. It generates an update gate vector and a candidate cell state vector using:
\begin{equation}
\bm{i}_t=\sigma(\mathbf{W}_{hi} \bm{h}_{t-1}+\mathbf{W}_{zi} \bm{z}_t+\bm{b}_i)
\end{equation}
\begin{equation}
\widetilde{\bm{C}}_t=\tanh(\mathbf{W}_{hc} \bm{h}_{t-1}+\mathbf{W}_{zc} \bm{z}_t+\bm{b}_c)
\end{equation}
where $\{ \mathbf{W}_{hi}, \mathbf{W}_{zi}, \bm{b}_i \}$ and $\{ \mathbf{W}_{hc}, \mathbf{W}_{zc}, \bm{b}_c \}$ denote recurrent weights, the input weights, and biases associated with the input and cell states, respectively. And $\tanh(.)$ is the hyperbolic tangent activation function. 
The cell state is then updated based on two terms, expressed as
\begin{equation}
\bm{C}_t=\bm{f}_t \odot \bm{C}_{t-1}+\bm{i}_t \odot \widetilde{\bm{C}}_t
\end{equation}
Here, the information from the previous state $\bm{C}_{t-1}$ is scaled by the forget gate output $\bm{f}_t$, while the candidate cell state $\widetilde{\bm{C}}_t$ is scaled by the input gate output $\bm{i}_t$ and added to form the updated cell state.
Subsequently, the output gate determines which part of the cell state contributes to the output. The hidden state $\bm{h}_t$ is computed by applying the hyperbolic tangent activation to the updated cell state $\bm{C}_t$, modulated by the output gate output $\bm{o}_t$, which is calculated as follows:
\begin{equation}
\bm{o}_t=\sigma(\mathbf{W}_{ho} \bm{h}_{t-1}+\mathbf{W}_{zo} \bm{z}_t+\bm{b}_o)
\end{equation}
\begin{equation}
\bm{h}_t=\bm{o}_t \odot \tanh(\bm{C}_t)
\end{equation}
where $\bm{o}_t$ represents the results of the output gate, and $\mathbf{W}_{ho}$, $\mathbf{W}_{zo}$, $\bm{b}_o$ represent the recurrent weight matrix, input weight matrix, and bias vector of the output gate, respectively. In regression tasks, the output $\bm{h}_t$ is connected to the training labels using a linear activation function. Gradients corresponding to the network's weights and biases can be derived based on the aforementioned LSTM architecture.

LSTM networks effectively capture temporal dependencies in time-series data. However, a key limitation lies in their inability to account for time-independent random variables during training, making it challenging to address reliability problems involving both time-dependent and time-independent uncertainties.

\subsection{DDF-LSTM Modeling Method}
\label{sect_DDF-LSTM}
\begin{figure*}
\centerline{\includegraphics[width=5in]{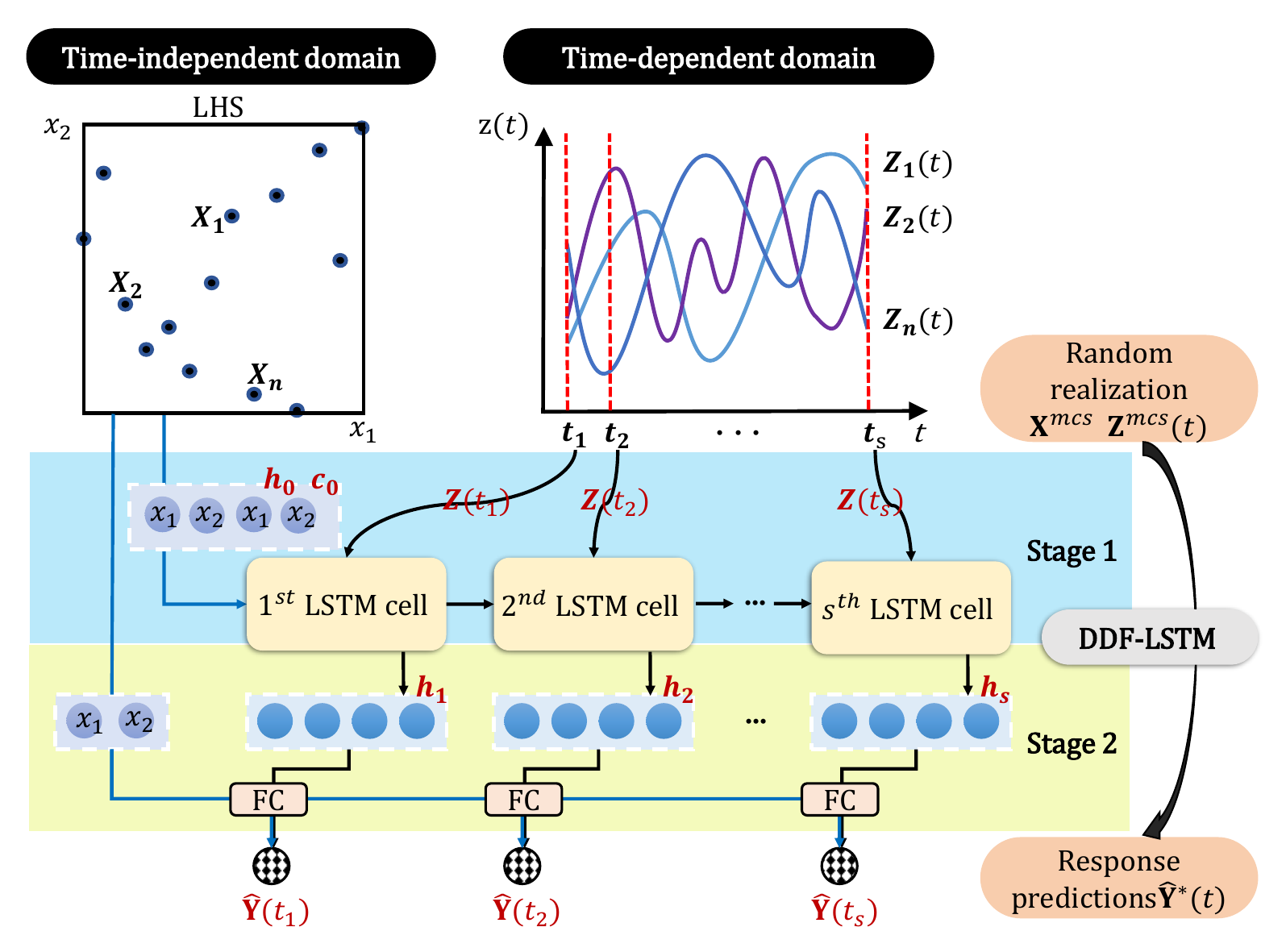}}
\caption{Sketch of the DDF-LSTM model architecture}
\label{fig2}
\end{figure*}

To address the limitations of LSTM in handling time-independent random variables, this study proposes a DDF-LSTM modeling method. As shown in Fig.~\ref{fig2}, the establishment of DDF-LSTM model is implemented through a two-stage strategy. In the first stage, time-independent random variables are embedded into the initial hidden states of the LSTM, replacing the conventional zero-initialization so that time-independent domain information is integrated from the very beginning. In the second stage, the outputs of the LSTM hidden states are processed through a fully-connected layer that fuses the time-independent and time-dependent features, yielding the final response prediction.

To prepare the training data, $n$ random variables $\mathbf{X}=[\bm{X}_1, \bm{X}_2, \dots, \bm{X}_n]$ are first generated through Latin Hypercube Sampling (LHS), which ensures coverage of the uniform range in the $d$-dimensional time-independent domain, effectively reducing sample redundancy and enhancing the representativeness of the samples. According to the stochastic properties, the corresponding time-dependent stochastic processes $\mathbf{Z}=[\bm{Z}_1(t), \bm{Z}_2(t), \dots, \bm{Z}_n(t)]$ are generated. For the $i$th realization, the time-dependent system response $\bm{y}_i$ is computed based on the input $[\bm{X}_i, \bm{Z}_i(t), t]$ as follows
\begin{equation}
\bm{y}_i=G[\bm{X}_i, \bm{Z}_i(t), t], \quad i=1,2,\dots,n
\end{equation}

To circumvent the memory redundancy caused by explicitly concatenating time-independent variables with time-dependent inputs, the proposed method embeds these static parameters into the LSTM's initial states, which serve as the starting conditions for processing sequential data. Instead of the conventional zero-vector initialization, these initial states are adapted based on the features of time-independent variables, the foundation for state updates throughout the sequence. 
To initialize the LSTM model, the hidden state $\bm{h}_0$ and the cell state $\bm{c}_0$ are set based on the feature vector $\bm{X}_i$ and the structure of LSTM.
The entire time interval is divided into $s$ nodes. The time-dependent data are structured into an input matrix, where the $i$th sample $\mathbf{M}_i$ is defined as
\begin{equation}
\mathbf{M}_i = 
\begin{bmatrix} \bm{m}_{i}^1 \\ \bm{m}_{i}^2 \\ \vdots \\ \bm{m}_{i}^s 
\end{bmatrix} = 
\begin{bmatrix} \bm{Z}_i(t_1) & t_1 \\ \bm{Z}_i(t_2) & t_2 \\ \vdots & \vdots \\ \bm{Z}_i(t_s) & t_s 
\end{bmatrix}, \quad i=1,2,\dots,n
\end{equation}
where $\bm{Z}_i(t_j)$ denotes the realization of the $i$th stochastic process at the $j$th time instant. A fully connected layer is appended to the LSTM hidden state outputs to further integrate time-independent information. This layer fuses the temporal hidden states $\bm{h}_i$ with the time-independent variables $\bm{x}_i$, mapping them to the final output space. The time-dependent response prediction for the $i$th sample at the $j$th time point is then formulated as:
\begin{equation}
\hat{y}_i(t_j)=f_m(\bm{h}_{i}^j, \bm{x}_i)
\end{equation}
where $f_m (.)$ represents the mapping function implemented by the fully connected layer, and $\bm{h}_{i}^j$ represents the LSTM hidden state outputs for the $i$th sample at the $j$th time step. The prediction matrix $\widehat{\mathbf{Y}}$, consisting of all samples and time steps, is expressed as:
\begin{equation}
\widehat{\mathbf{Y}}=[\hat{\bm{y}}_1, \hat{\bm{y}}_2, \dots, \hat{\bm{y}}_n] = \begin{bmatrix} \hat{y}_1(t_1) & \cdots & \hat{y}_n(t_1) \\ \vdots & \ddots & \vdots \\ \hat{y}_1(t_s) & \cdots & \hat{y}_n(t_s) \end{bmatrix}
\end{equation}

Time-dependent reliability assessment emphasizes the minimum response across the entire time interval, which serves as a key indicator for system failure. Therefore, surrogate models should focus more on the capability of accurately estimating the minimum response in terms of ensuring precise reliability approximation. To improve the sensitivity of the DDF-LSTM model on the minimum response, a novel loss function is designed as
\begin{equation}
f_{loss}=\alpha/(1+\bm{A}) * MSE(\hat{\bm{y}}, \bm{y})
\label{eq_loss}
\end{equation}
where $\bm{A} = |\bm{y} - y_{min}| / |y_{max} - y_{min}|$. The Mean Squared Error (MSE) quantifies prediction accuracy by averaging the squared deviations between predicted values $\hat{\bm{y}}$ and actual values $\bm{y}$, and is commonly adopted as a standard performance metric. For each sample, $y_{min}$ and $y_{max}$ represent the minimum and maximum values extracted from its time sequence corresponding, respectively. The term $\bm{A}$ quantifies the normalized distance of the response at each time step from the sample-specific minimum, enabling the loss scaling to be independent of the global response range. The factor term $\alpha/(1+\bm{A})$ increases when the response approaches the minimum response, leading to more weights to the loss. The hyperparameter $\alpha$ controls the emphasis placed on regions near the minimum response. A value of $\alpha = 10$ was selected for all case studies in this work. This choice was empirically found to provide a strong and consistent weighting factor that effectively amplifies the loss around the minimum response—which is critical for accurate reliability estimation—without overwhelming the gradient signal from the rest of the sequence. For applications with different objectives, $\alpha$ can be treated as a tunable parameter. Compared with the conventional MSE loss, this novel loss function effectively enhances the model's ability to characterize extreme responses, thereby enhancing time-dependent reliability prediction accuracy.

\subsection{Time-Dependent Reliability Analysis Using DDF-LSTM}
\label{sect_procedure}
\begin{figure}
\centerline{\includegraphics[width=\columnwidth]{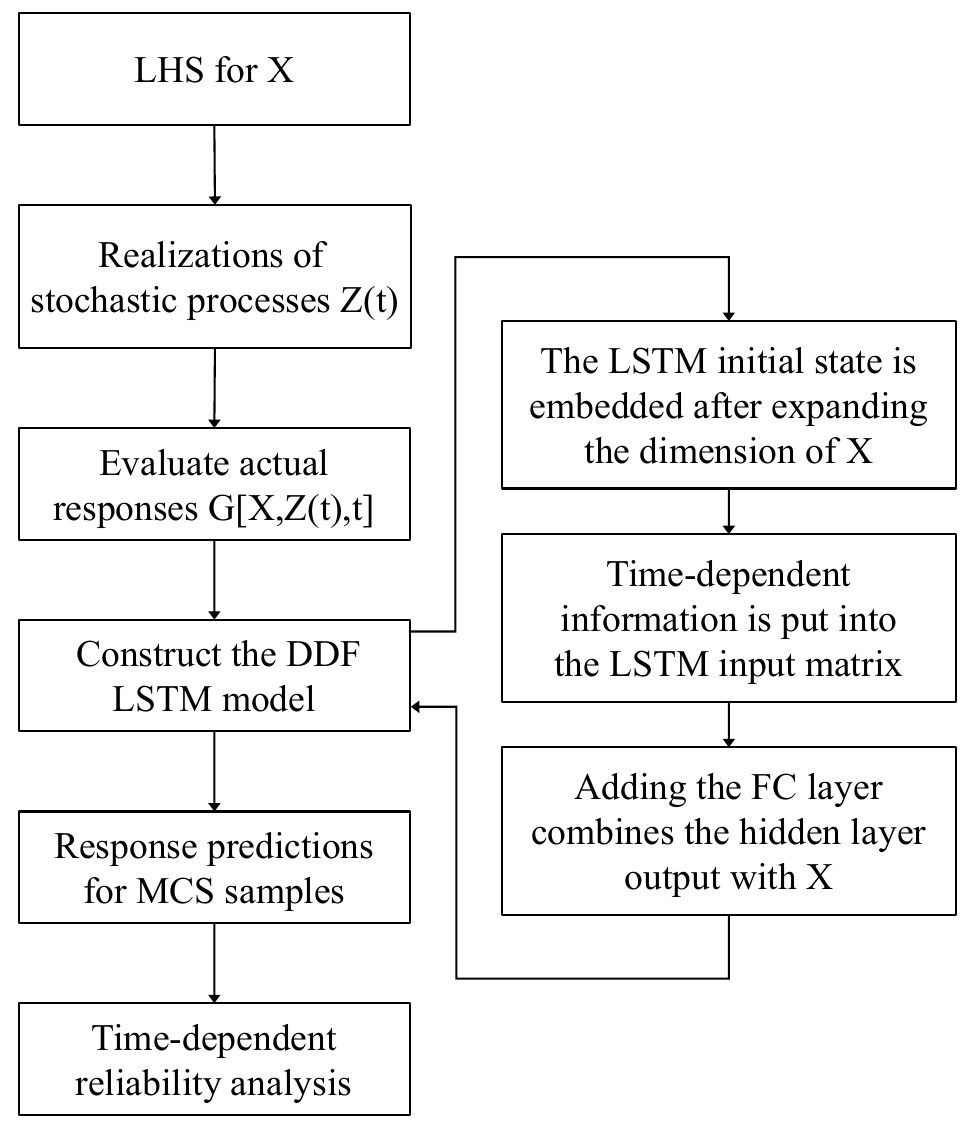}}
\caption{Procedure of the DDF-LSTM framework for time-dependent reliability analysis}
\label{fig3}
\end{figure}

The overall procedure of the proposed DDF-LSTM framework for efficient time-dependent reliability analysis is depicted in Fig.~\ref{fig3}. Initially, $n$ samples of the time-independent input parameters $\mathbf{X}$ are generated using the LHS method. Correspondingly, $n$ samples of the stochastic process $\mathbf{Z}(t)$ are constructed based on their statistical properties using the EOLE method. Following the procedure described in section~\ref{sect_DDF-LSTM}, in order to construct the proposed DDF-LSTM model, the time-independent information is embedded into the initial hidden state $\bm{h}_0$ and cell state $\bm{c}_0$ of the LSTM. The actual time-dependent responses $\bm{y}_i(t)$, $i = 1, ..., n$, serve as the training labels for the DDF-LSTM model. The loss function presented in Eqn.~(\ref{eq_loss}) is adopted to minimize the discrepancy between the training labels and the predictions, and determine the values of the weights and biases effectively. The proposed method is implemented in a Python 3.8 environment, where the Adam optimizer is used to train the model with a default learning rate of 0.001. To ensure computational efficiency during network training, standardization is performed for data pre-processing.

To conduct time-dependent reliability analysis, $N$ MCS samples are generated according to the statistical information of input variables and stochastic processes, denoted as $\mathbf{U}^{mcs}=[\mathbf{X}^{mcs}, \mathbf{Z}^{mcs}(t)]$, where $\mathbf{X}^{mcs}=[\bm{X}_1^m, \bm{X}_2^m, \dots, \bm{X}_N^m]$, and $\mathbf{Z}^{mcs}(t)=[\bm{Z}_1^m(t), \bm{Z}_2^m(t), \dots, \bm{Z}_N^m(t)]$. Using the trained model, accurate predictions of the time-dependent system response $G(\mathbf{X}^{mcs}, \mathbf{Z}^{mcs}(t), t)$ can be achieved by providing the inputs $\mathbf{U}^{mcs}$, expressed as
\begin{equation}
\widehat{\mathbf{Y}}^*= [\hat{\bm{y}}_1^*, \hat{\bm{y}}_2^*, \dots, \hat{\bm{y}}_N^*] = 
\begin{bmatrix} \hat{y}_1^*(t_1) & \cdots & \hat{y}_N^*(t_1) \\ \vdots & \ddots & \vdots \\ \hat{y}_1^*(t_s) & \cdots & \hat{y}_N^*(t_s) 
\end{bmatrix}
\end{equation}
After the predicting system responses, the minimum value per MCS sample is extracted and utilized for time-dependent reliability estimation. The failure event is then identified based on the following indicator function:
\begin{equation}
I_f(\hat{\bm{y}}_i^*)=
\begin{cases} 
1, & \min_{t_1 \leq t_j \leq t_s} \hat{y}_i^*(t_j) < 0 \\
0, & \text{otherwise}
\end{cases}
\label{eq_If_function}
\end{equation}
where $\hat{y}_i^*(t_j)$ denotes the predicted time-dependent response of the $i$th MCS sample at time instant $t_j$. A failure is considered to occur if the minimum response over the interval $[0,T]$ falls below zero. The time-dependent failure probability $P_f$ is estimated as follows:
\begin{equation}
P_f(0,T) \approx \frac{N_f}{N}
\end{equation}
where $N_f$ is the number of samples classified as failures, and $N$ is the total number of MCS samples.

In addition, a training strategy that utilizes a subset of time interval instead of the full-time interval is explored in the case study section. This approach allows for the assessment of the model's generalization ability under limited temporal information. Even with reduced training data, the proposed DDF-LSTM model maintains strong predictive performance, thereby confirming its robustness and efficiency in time-dependent reliability analysis.

\section{Case studies}
\label{sect_case}

Four case studies are conducted in this section to validate the performance and advantages of the proposed method for time-dependent reliability assessment.

\subsection{Case Study I: Multivariate Mathematical Example}
This example involves a multivariate mathematical problem in which the limit state function $G(\mathbf{X},\mathbf{Z}(t))$ consists of three random variables and five stochastic processes, expressed as 
\begin{equation}
    \begin{split}
        G(\mathbf{X},\mathbf{Z}(t)) &= 0.5x_1^2z_2(t)z_3(t)-8z_1(t)z_2(t) \\
        &+(x_2+1)^2+x_3^2z_4(t)-5z_5(t)-20
    \end{split}
    \label{eq_G1}
\end{equation}
Among them, each time-independent random variable in $\mathbf{X}=[x_1, x_2, x_3]$ is assumed to follow a normal distribution, while each time-dependent stochastic process is modeled as a stationary Gaussian process. The time interval [0, 1] is uniformly partitioned into 60 discrete time nodes, and the autocorrelation function for the $i$th stochastic process can be formulated as
\begin{equation}
\rho_i(t_1, t_2)=\exp\left(-\frac{(t_2-t_1)^2}{\lambda_i}\right)
\end{equation}
where $\lambda_i$ are assigned to be 0.01, 0.005, 0.005, 0.002, and 0.003, respectively. Detailed statistical properties of the input variables are listed in Table~\ref{tab_variable1}.
\begin{table}
\caption{Statistical characteristics of the random variables and stochastic processes}
\label{tab_variable1}
\begin{center}
\small  
\begin{tabular}{c c c c}
\hline
Input variable & Distribution & Mean & \makecell{Standard\\ deviation} \\
\hline
$x_1$ & Normal & 5 & 0.5 \\
$x_2$ & Normal & 6 & 0.5 \\
$x_3$ & Normal & 3.5 & 0.3 \\
$z_1(t)$ & Stationary Gaussian & 5 & 0.3 \\
$z_2(t)$ & Stationary Gaussian & 2 & 0.1 \\
$z_3(t)$ & Stationary Gaussian & 4 & 0.2 \\
$z_4(t)$ & Stationary Gaussian & 4 & 0.1 \\
$z_5(t)$ & Stationary Gaussian & 8 & 0.25 \\
\hline
\end{tabular}
\end{center}
\end{table}

To implement the proposed approach, the initial step involves constructing the training dataset for the DDF-LSTM model. 15 samples of random variables $\mathbf{X}$ and stochastic processes $\mathbf{Z}(t)$ are generated, and the system responses are subsequently computed according to Eqn.~(\ref{eq_G1}). 
Following Section~\ref{sect_DDF-LSTM}'s methodology, a DDF-LSTM model is constructed based on the 15 training samples, with 15 neurons set in the hidden layer. The time-independent input information is encoded in the initial LSTM states, and the hidden layer output at each time step is fused with the time-independent information to obtain the target value. The training process ends once the training loss falls below $10^{-4}$. Upon convergence, the trained DDF-LSTM model can be directly applied to predict system responses for new input realizations.

\begin{figure*}
\centering
    
    \begin{subfigure}[b]{\columnwidth}
        \centering
        \includegraphics[width=\linewidth]{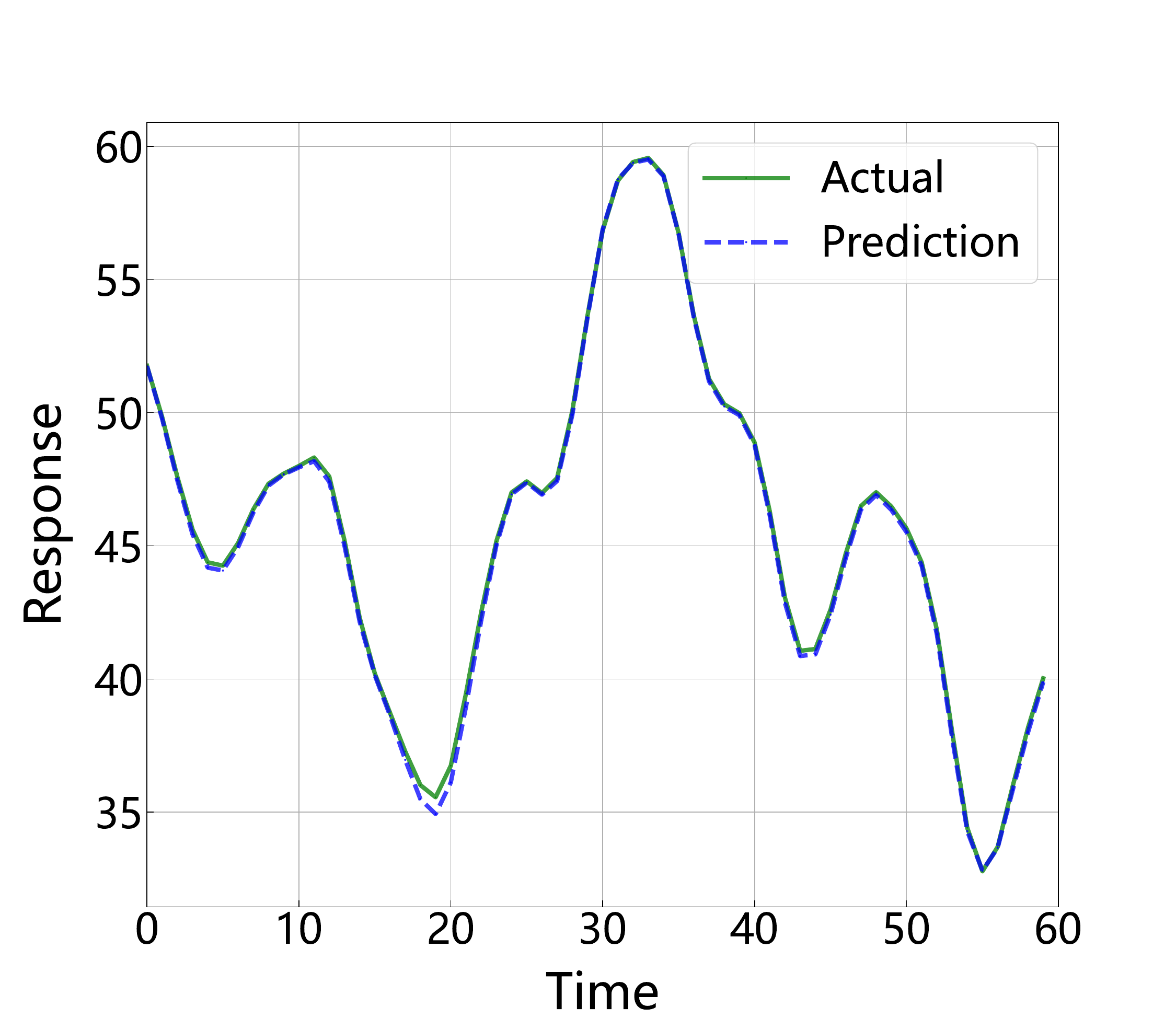}
        \caption{}
    \end{subfigure}
    \begin{subfigure}[b]{\columnwidth}
        \centering
        \includegraphics[width=\linewidth]{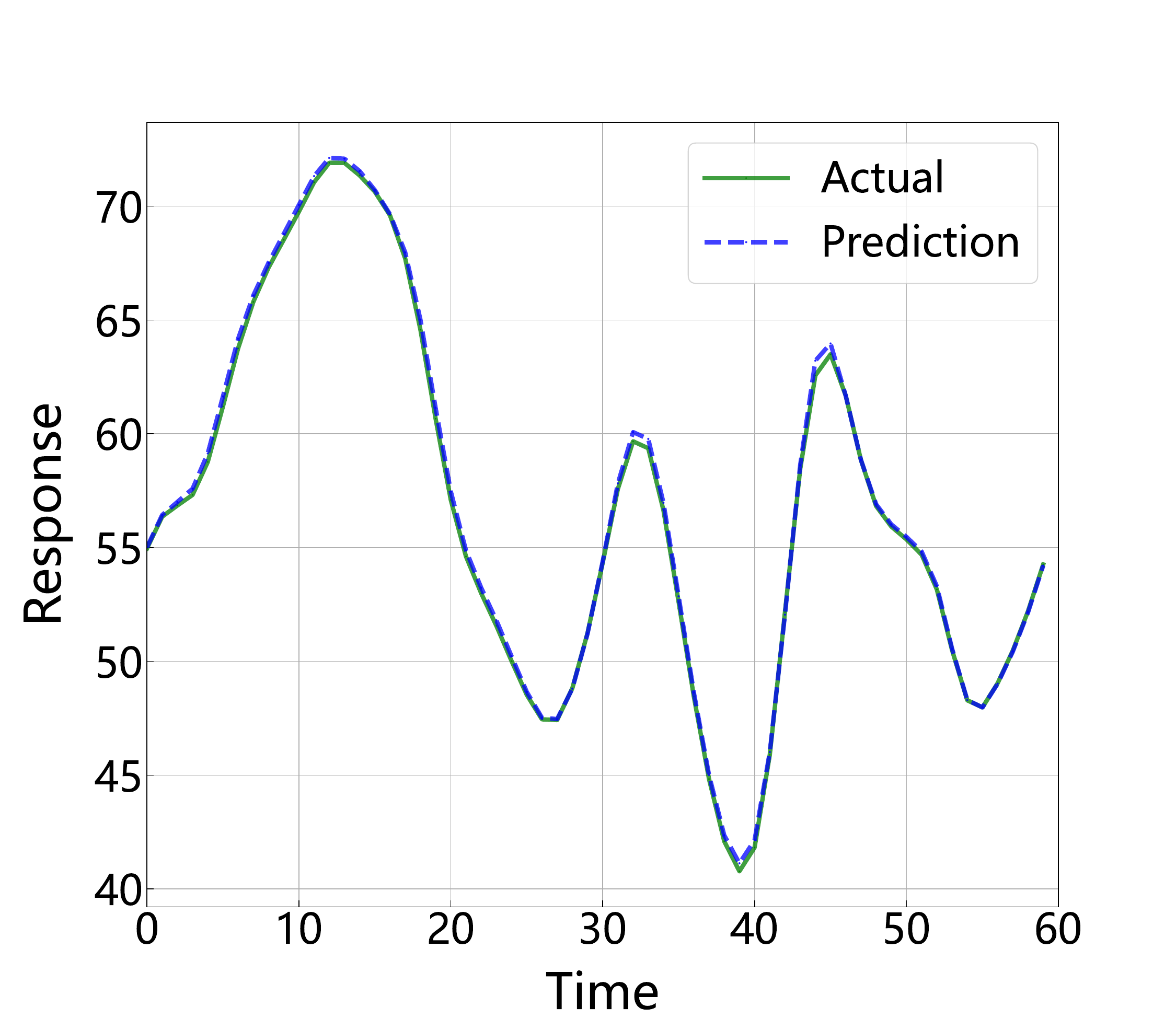}
        \caption{}
    \end{subfigure}

\caption{Comparison of DDF-LSTM predicted and actual responses for MCS samples: (a) 1st and (b) 75th}
\label{fig5}
\end{figure*}

To estimate system reliability, $10^6$ MCS samples are generated and the corresponding response predictions are obtained using the DDF-LSTM model. For demonstrating the accuracy of the DDF-LSTM model in capturing the system dynamics, Fig.~\ref{fig5} compares the predicted and actual responses for the 1st and 75th MCS samples in subfigures (a) and (b), respectively. The predicted curves exhibit near-complete overlap with the true responses, indicating excellent global accuracy. These results demonstrate the DDF-LSTM model's effectiveness in handling time-dependent uncertainty and its capability to accurately predict the system responses for the entire time series.

\begin{figure}
\centerline{\includegraphics[width=\columnwidth]{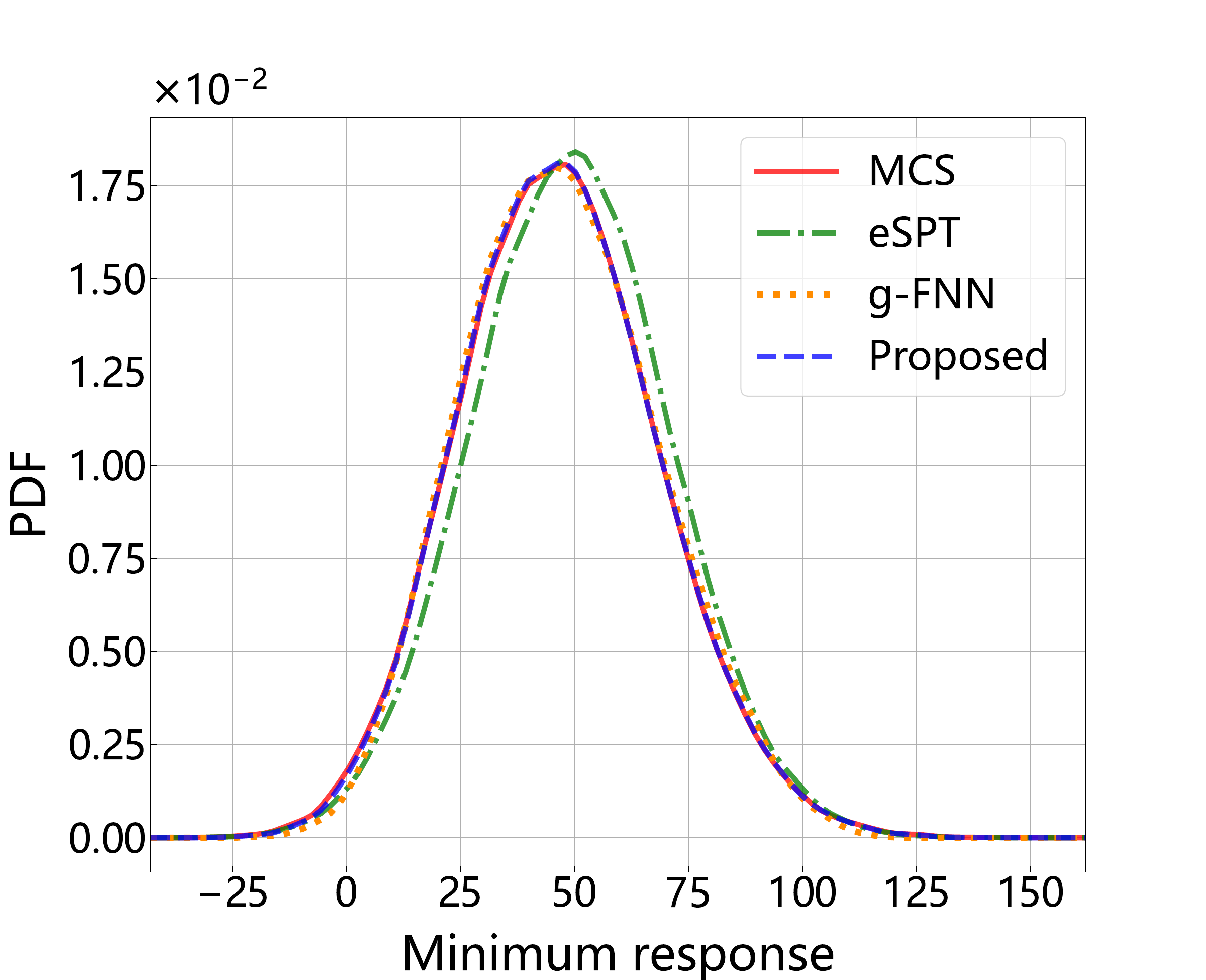}}
\caption{Comparison of the PDF of minimum responses using different methods}
\label{fig6}
\end{figure}

The actual time-dependent responses of $10^6$ MCS samples are evaluated, yielding a reference reliability of 0.98743. Based on the minimum response values, the MCS samples are classified as either failure or safe using Eqn.~(\ref{eq_If_function}), resulting in a reliability estimate of 0.98897. In order to evaluate the advantages of the proposed approach, two existing approaches—the eSPT method \cite{wang2016time} (with a confidence level of 0.999) and the g-FNN method \cite{li2022lstm}—are employed for comparison. The eSPT method transforms the random process and the time parameters into random parameters, and constructs a time-independent reliability model that covers all potential failure events throughout the entire time interval. The g-FNN method first employs LSTM to train local surrogate models and subsequently generates augmented datasets using multiple LSTMs to train a global FNN surrogate model. Fig.~\ref{fig6} presents a comparison of the minimum response PDF calculated by different methods. The results indicate that the minimum response PDF obtained by the proposed approach has the best consistency with the one achieved by MCS. Table~\ref{tab_results1} summarizes the reliability estimations obtained by different methods. Though these methods can provide accurate reliability approximations, eSPT methods require 23 time series, whereas the proposed method requires only 15 time series with higher accuracy. It can be seen that the relative error of reliability obtained by g-FNN is slightly larger than that of the proposed method. The reason lies in that the multiple network structures employed in the g-FNN method increase model uncertainty, which in turn affects the accuracy of the final prediction. Furthermore, this method requires training multiple LSTM networks based on fixed time-independent variables, resulting in longer training times compared to the proposed approach. In summary, the proposed method unifies both time-dependent and time-independent information within a dual-domain fused modeling framework. By maintaining high prediction accuracy, it substantially reduces the required amount of time-series data and shortens the training duration. This balance of computational efficiency and predictive performance makes the approach a more competitive solution for practical time-dependent reliability analysis.

\begin{table}
\caption{Comparison results}
\begin{center}
\label{tab_results1}
\resizebox{\columnwidth}{!}{%
\begin{tabular}{l c c c}
\hline
Method & \makecell{No. of Time-\\series data} & Reliability & Relative error (\%) \\
\hline
MCS & $10^6$ & 0.98743 & — \\
eSPT & $\sim$23 & 0.99007 & 0.26736 \\
g-FNN & 15 & 0.99339 & 0.60359 \\
Proposed & 15 & 0.98897 & 0.15596 \\
\hline
\end{tabular}
}%
\end{center}
\end{table}

To verify the robustness capability of the DDF-LSTM model, time-dependent reliability evaluations are conducted on several modified problem settings. Table~\ref{tab_robustness1} summarizes the configuration details of three representative scenarios, each employing three distinct training datasets. The proposed method consistently delivers stable and accurate reliability estimates across all scenarios, with minimal relative errors observed. These results indicate the method's effectiveness and robustness in addressing time-dependent reliability problems with multivariate stochastic processes.

\begin{table}
\caption{Results of robustness testing in the mathematical case study}
\begin{center}
\label{tab_robustness1}
\resizebox{\columnwidth}{!}{%
\begin{tabular}{c c c c c c}
\hline
Scenarios & Modification & Sample & MCS rel. & Est. rel. & Rel. error (\%) \\
\hline
1 & — & 1) & 0.98743 & 0.98618 & 0.12659 \\
 & & 2) & 0.98743 & 0.99020 & 0.28053 \\
 & & 3) & 0.98743 & 0.98668 & 0.07595 \\
 & & Average & 0.98743 & 0.98769 & 0.16102 \\
2 & $x_1 \sim N(4.3,0.4)$ & 1) & 0.89596 & 0.89369 & 0.25336 \\
 & & 2) & 0.89596 & 0.89383 & 0.23773 \\
 & & 3) & 0.89596 & 0.89162 & 0.48440 \\
 & & Average & 0.89596 & 0.89305 & 0.32516 \\
3 & $\mu(z_1)=4.2$ & 1) & 0.99581 & 0.99772 & 0.19180 \\
 & $\sigma(z_1)=0.5$ & 2) & 0.99581 & 0.99862 & 0.28218 \\
 & & 3) & 0.99581 & 0.99639 & 0.05824 \\
 & & Average & 0.99581 & 0.99758 & 0.17741 \\
\hline
\end{tabular}
}
\end{center}
\end{table}

\begin{figure}
\centerline{\includegraphics[width=\columnwidth]{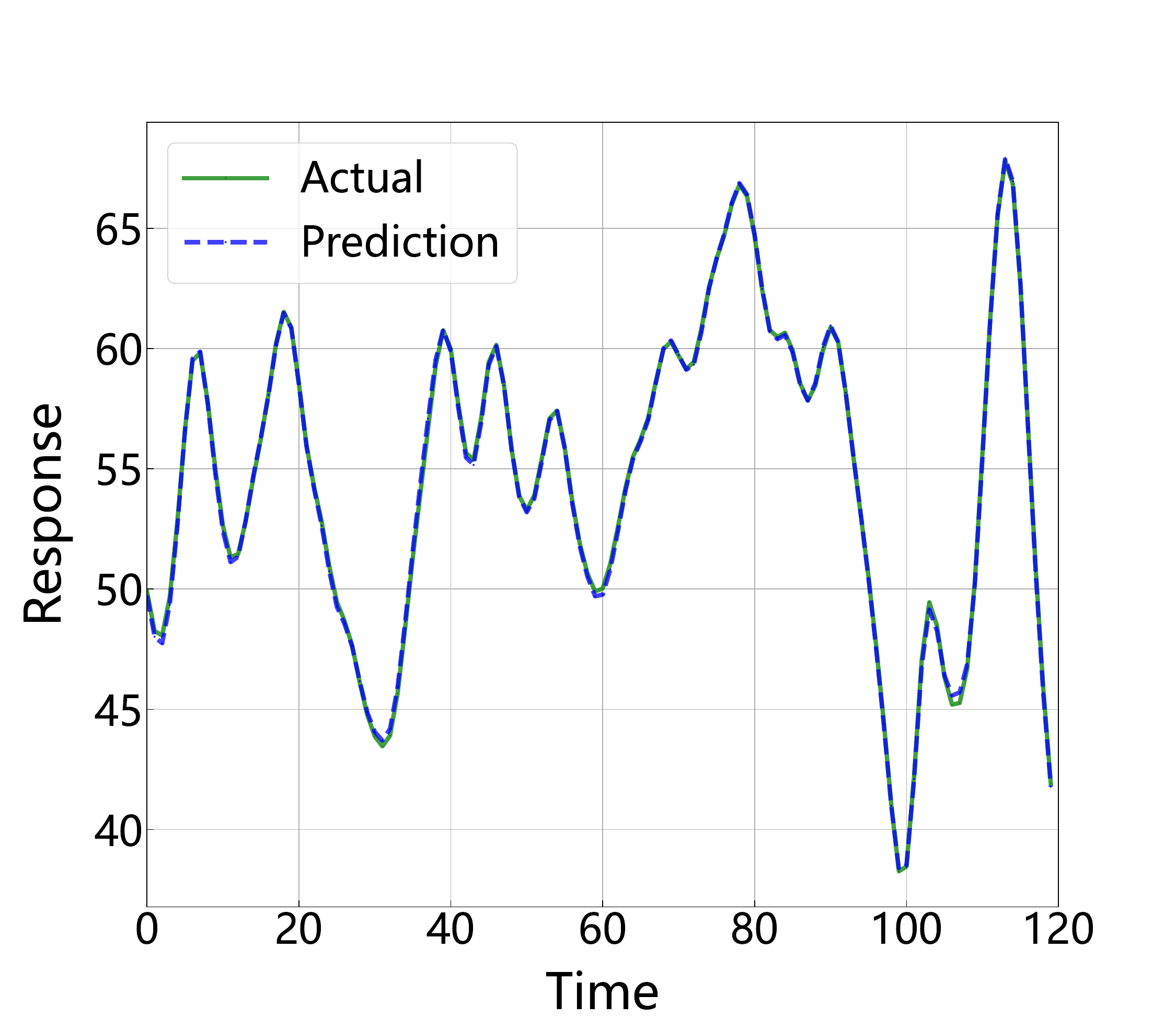}}
\caption{Comparison between actual and predicted responses in 120 time intervals}
\label{fig7}
\end{figure}

\begin{figure}
\centerline{\includegraphics[width=\columnwidth]{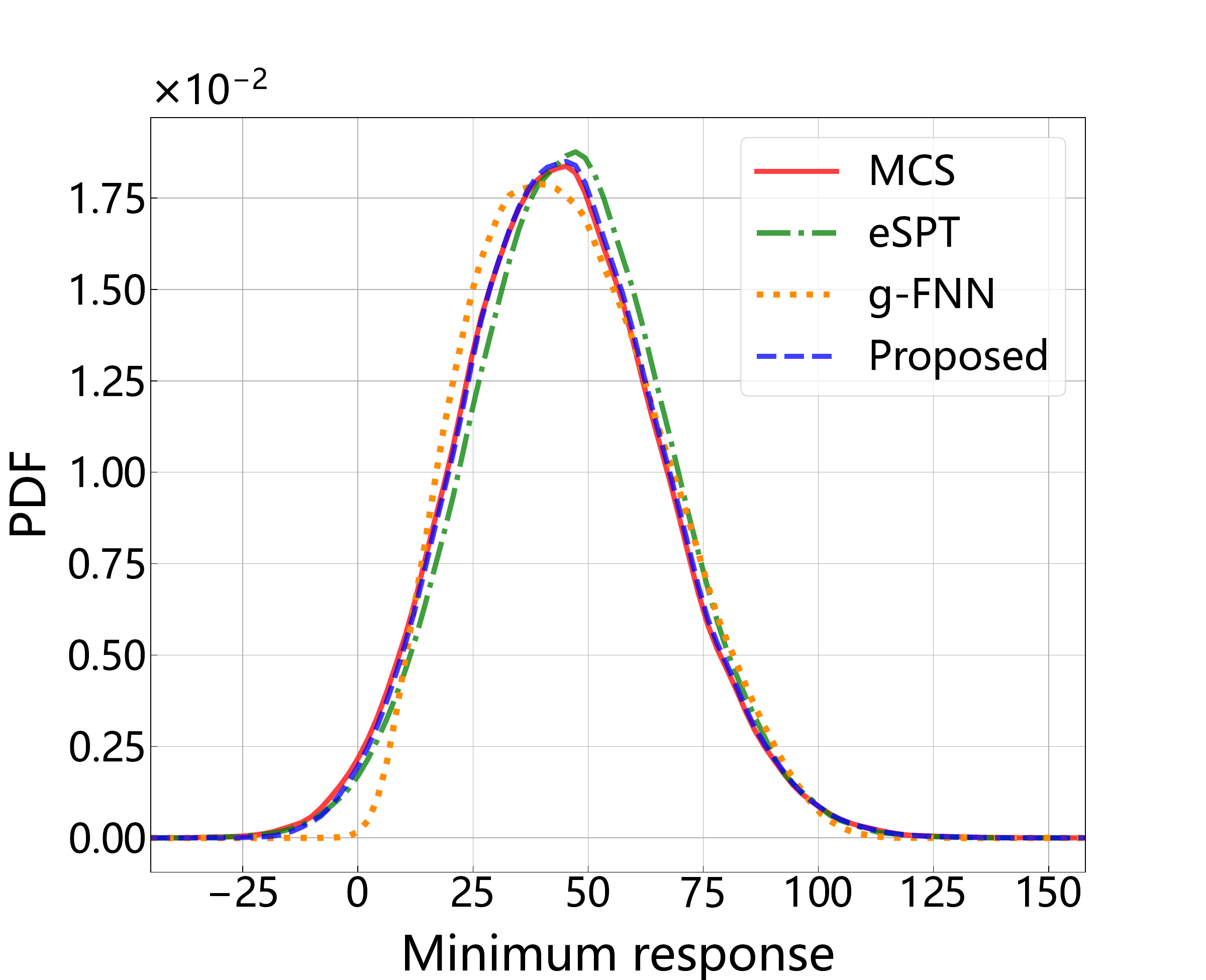}}
\caption{PDF comparison for all MCS samples in 120 time intervals}
\label{fig8}
\end{figure}

To evaluate the proposed model's capability in predicting system responses beyond the training range, the future time interval is extended from 60 to 120 time nodes while maintaining the original temporal step size. Meanwhile, the training sequences are generated with only the first 60 time nodes, leaving the subsequent 60 nodes unseen during training. Fig.~\ref{fig7} presents a comparison between the true system response and the DDF-LSTM prediction for the 1st MCS sample over the 120-step interval. The results indicate that the DDF-LSTM model retains a close agreement with the actual responses across the extended prediction period. For comparison purpose, Fig.~\ref{fig8} and Table~\ref{table4} summarize the reliability analysis results of different methods. In the case of the eSPT method, time-dependent stochastic processes are transformed into time-independent random parameters. Moreover, since the limit state function does not explicitly depend on time $t$, reducing the length of training sequences does not affect the performance of eSPT, and its sample requirement remains unchanged. For consistency, the g-FNN method is trained using 15 time series, each comprising only the first 60 time nodes. It is worth noting that this case involves five stochastic processes, whose number and length directly determine the input dimension of the g-FNN model, thereby increasing computational time and causing potential instability. In contrast, the proposed DDF-LSTM approach, achieves a reliability estimate of 0.98783 with only 0.40964\% relative error. These results highlight the superior accuracy and scalability of the proposed method for long-term sequence prediction in time-dependent reliability analysis.

\begin{table}
\caption{Comparison results Comparison results in 120 time intervals}
\begin{center}
\label{table4}
\resizebox{\columnwidth}{!}{%
\begin{tabular}{l c c c}
\hline
Method & \makecell{No. of Time-\\series data} & Reliability & Relative error (\%) \\
\hline
MCS & $10^6$ & 0.98380 & — \\
eSPT & $\sim$23 & 0.98721 & 0.34662 \\
g-FNN & 15 & 0.99985 & 1.63143 \\
Proposed & 15 & 0.98783 & 0.40964 \\
\hline
\end{tabular}
}
\end{center}
\end{table}

\subsection{Case Study II: A Corroded Beam Problem}
\begin{figure}
\centerline{\includegraphics[width=\columnwidth]{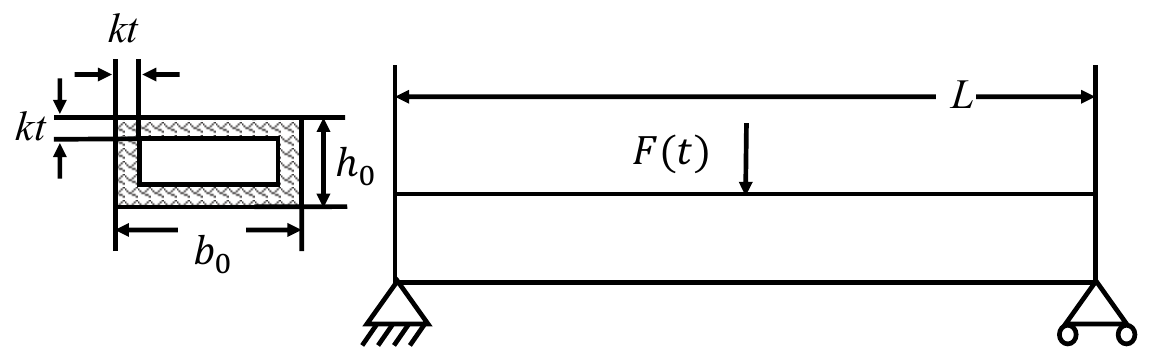}}
\caption{Schematic of the beam geometry and its corroded cross-section}
\label{fig9}
\end{figure}
A beam corrosion case is used to evaluate the performance of the proposed method. As shown in Fig.~\ref{fig9}, a stochastic concentrated load $F(t)$ is applied to the beam's midspan, which is modeled as a stationary Gaussian process. Initial cross-sectional dimensions of the beam are denoted as $b_0$ and $h_0$, and both dimensions degrade over time due to corrosion. The cross-section dimensions $b(t)$ and $h(t)$ over time can be modeled as time-dependent random variables:
\begin{equation}
\begin{cases} 
b(t)=b_0-2kt \\ 
h(t)=h_0-2kt
\end{cases}
\end{equation}
where $k=5 \times 10^{-5}\ \text{m/month}$ is the corrosion rate constant. The limit state function for this problem incorporates three random variables, a stochastic process, and a time parameter, denoted as:
\begin{equation}
    \begin{split}
        G(\mathbf{X},\mathbf{Z}(t),t) &= 0.25b(t)h(t)^2\sigma_y-0.25 F(t)L\\
                                      &-0.125\rho b(t)h(t)L^2
    \end{split}
    \label{eq_G2}
\end{equation}
Here, $\sigma_y$ denotes the yield strength of the material, $L = 5\ \text{m}$ is the beam length, $\rho=785\ \text{g/cm}^3$ represents the material density. The system is considered to fail once the maximum stress exceeds the yield strength. The analysis interval ranges from 1 to 30 months, discretized into 100 time points. Table~\ref{table5} summarizes the input distributions, and the autocorrelation of the random load $F(t)$ is defined as:
\begin{equation}
\rho_{F}(t_1,t_2)=\exp\left(-(t_1-t_2)^2\right)
\end{equation}

\begin{table}
\caption{Statistical properties of input variables in the corroded beam example}
\begin{center}
\label{table5}
\resizebox{\columnwidth}{!}{%
\begin{tabular}{l l l l}
\hline
Input variable & Distribution & Mean & \makecell{Standard \\deviation} \\
\hline
$\sigma_y$ (MPa) & Normal & 250 & 24 \\
$b_0$ (m) & Normal & 0.6 & 0.01 \\
$h_0$ (m) & Normal & 0.06 & 0.004 \\
$F(t)$ (N) & Stationary Gaussian & 3500 & 700 \\
\hline
\end{tabular}
}
\end{center}
\end{table}

\begin{figure}
\centerline{\includegraphics[width=\columnwidth]{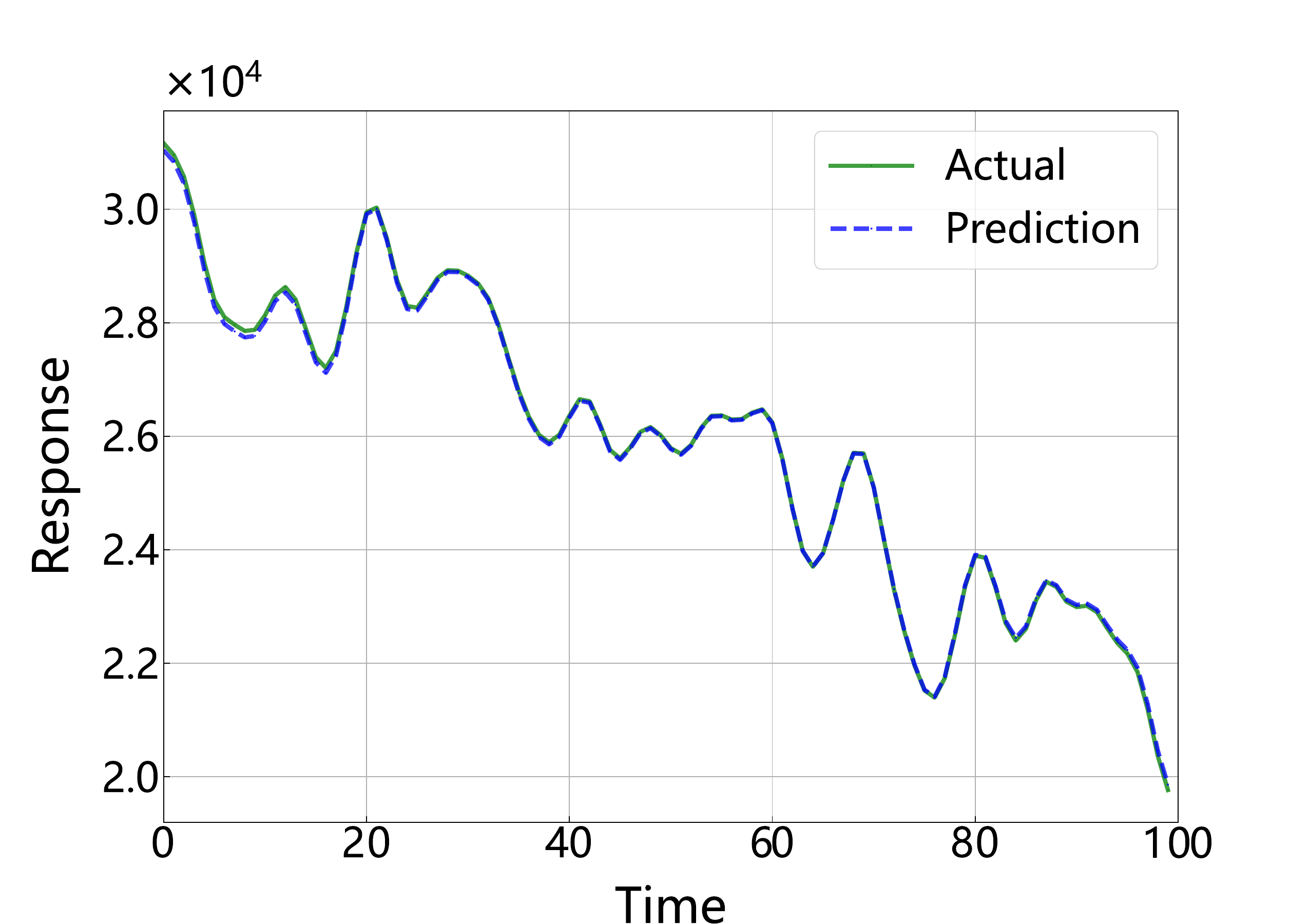}}
\caption{Comparison of DDF-LSTM predicted and actual responses}
\label{fig10}
\end{figure}

\begin{figure}
\centerline{\includegraphics[width=\columnwidth]{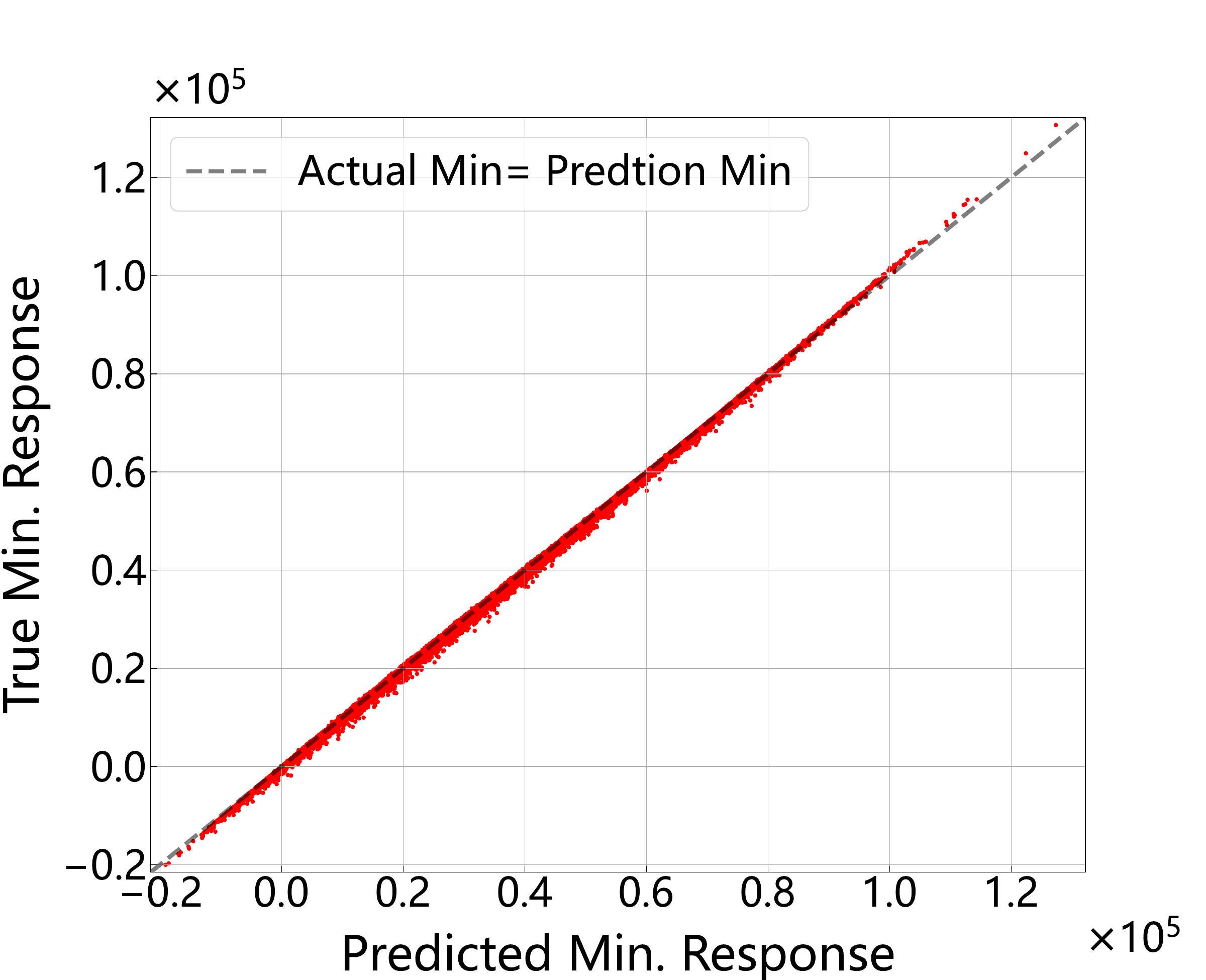}}
\caption{Minimum response predictions for all MCS samples}
\label{fig11}
\end{figure}

Building upon the procedure established in case study I, the DDF-LSTM model is trained using 12 datasets, with the number of hidden neurons set to 15. Utilizing the trained model, time-dependent responses for $10^6$ MCS samples are estimated. Fig.~\ref{fig10} presents a comparison between the actual and predicted responses for the 100th sample, demonstrating the model's accuracy in predicting time-dependent responses. Furthermore, Fig.~\ref{fig11} compares the actual and predicted minimum response values across all MCS samples. The predicted minimum responses show strong agreement with the actual values, demonstrating the model's ability to precisely forecast extreme responses throughout the entire time interval.

\begin{figure}
\centerline{\includegraphics[width=\columnwidth]{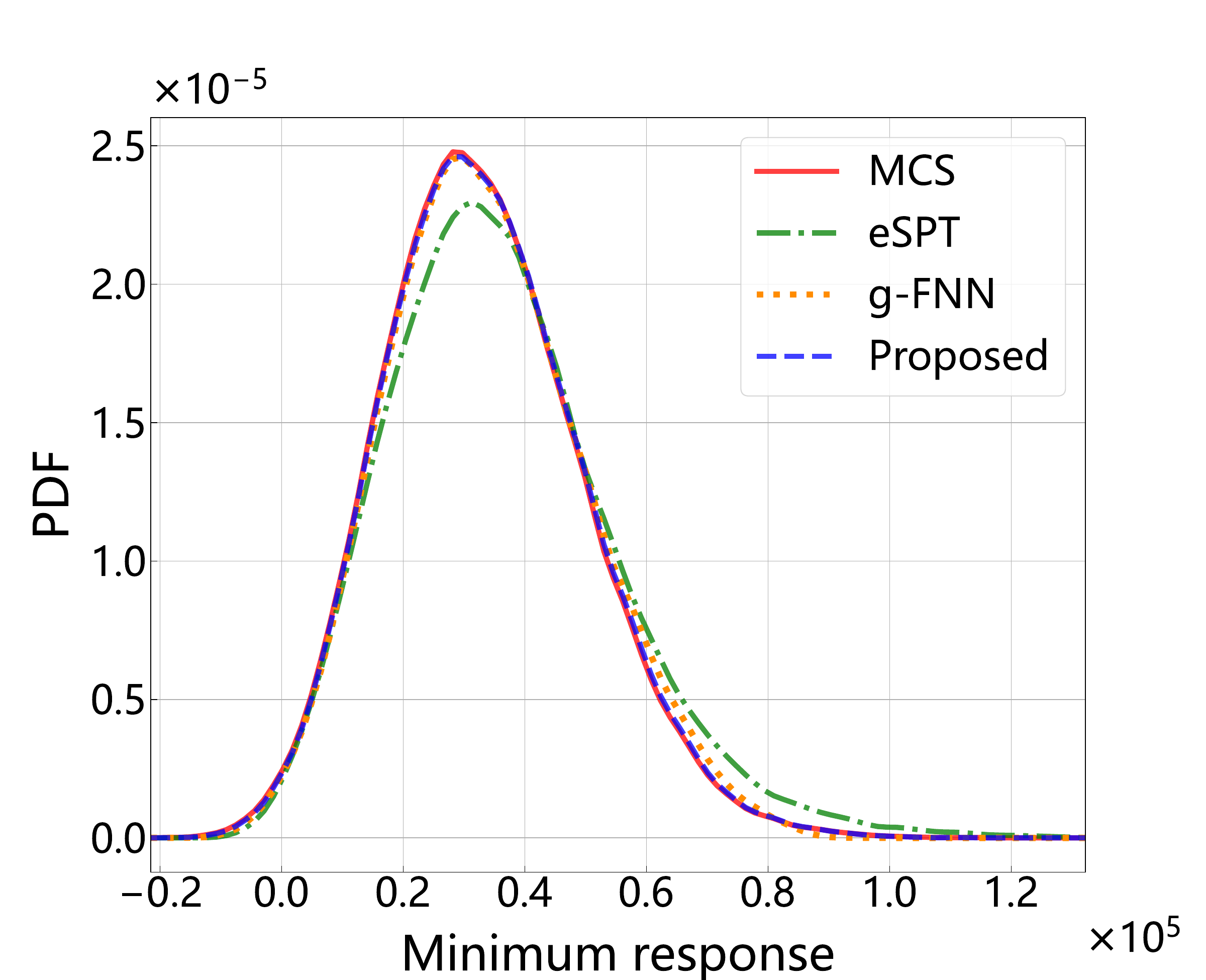}}
\caption{PDF of the minimum responses using different methods}
\label{fig12}
\end{figure}

By adjusting the prediction horizon, the DDF-LSTM model can be used to approximate time-dependent reliability over various subintervals $[0, T_i]$, $T_i \leq T$. Table~\ref{table6} presents the estimated reliability values within different time intervals, demonstrating that the proposed method achieves high accuracy when compared with reference values obtained from direct MCS. A comparison of different methods is provided in Table~\ref{table7} and Fig.~\ref{fig12}. Compared with other methods, the proposed approach provides comparable accuracy with substantially less time-series training data.

\begin{table}
\caption{Estimated time-dependent reliability over different time intervals}
\begin{center}
\label{table6}
\resizebox{\columnwidth}{!}{%
\begin{tabular}{c c c c}
\hline
Time nodes & Accurate RA & Estimated RA & Relative error (\%) \\
\hline
$t_1 \sim t_{20}$ & 0.99608 & 0.99655 & 0.04718 \\
$t_1 \sim t_{40}$ & 0.99501 & 0.99554 & 0.05327 \\
$t_1 \sim t_{60}$ & 0.99354 & 0.99425 & 0.07146 \\
$t_1 \sim t_{80}$ & 0.99173 & 0.99255 & 0.08268 \\
$t_1 \sim t_{100}$ & 0.98956 & 0.99042 & 0.08691 \\
\hline
\end{tabular}
}
\end{center}
\end{table}

\begin{table}
\caption{Comparison results for time interval [1, 30]}
\begin{center}
\label{table7}
\resizebox{\columnwidth}{!}{%
\begin{tabular}{l c c c}
\hline
Method & \makecell{No. of Time-\\series data} & Reliability & Relative error (\%) \\
\hline
MCS & $10^6$ & 0.98956 & — \\
eSPT & $\sim$93 & 0.99401 & 0.44969 \\
g-FNN & 12 & 0.99162 & 0.20817 \\
Proposed & 12 & 0.99042 & 0.08691 \\
\hline
\end{tabular}
}
\end{center}
\end{table}

\begin{figure}
\centerline{\includegraphics[width=\columnwidth]{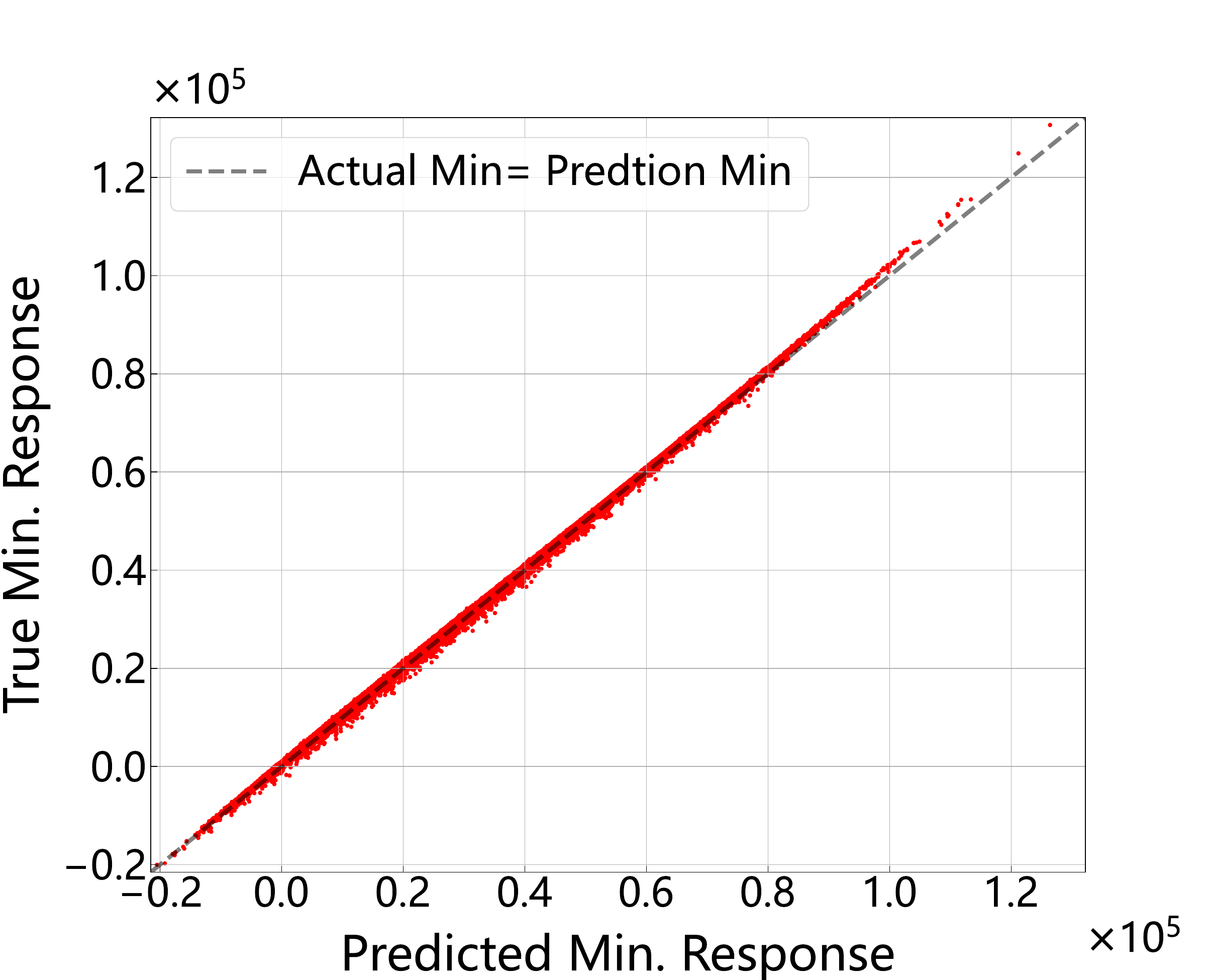}}
\caption{Predicted vs. actual minimum responses across all MCS samples using DDF-LSTM}
\label{fig14}
\end{figure}

\begin{figure}
\centerline{\includegraphics[width=\columnwidth]{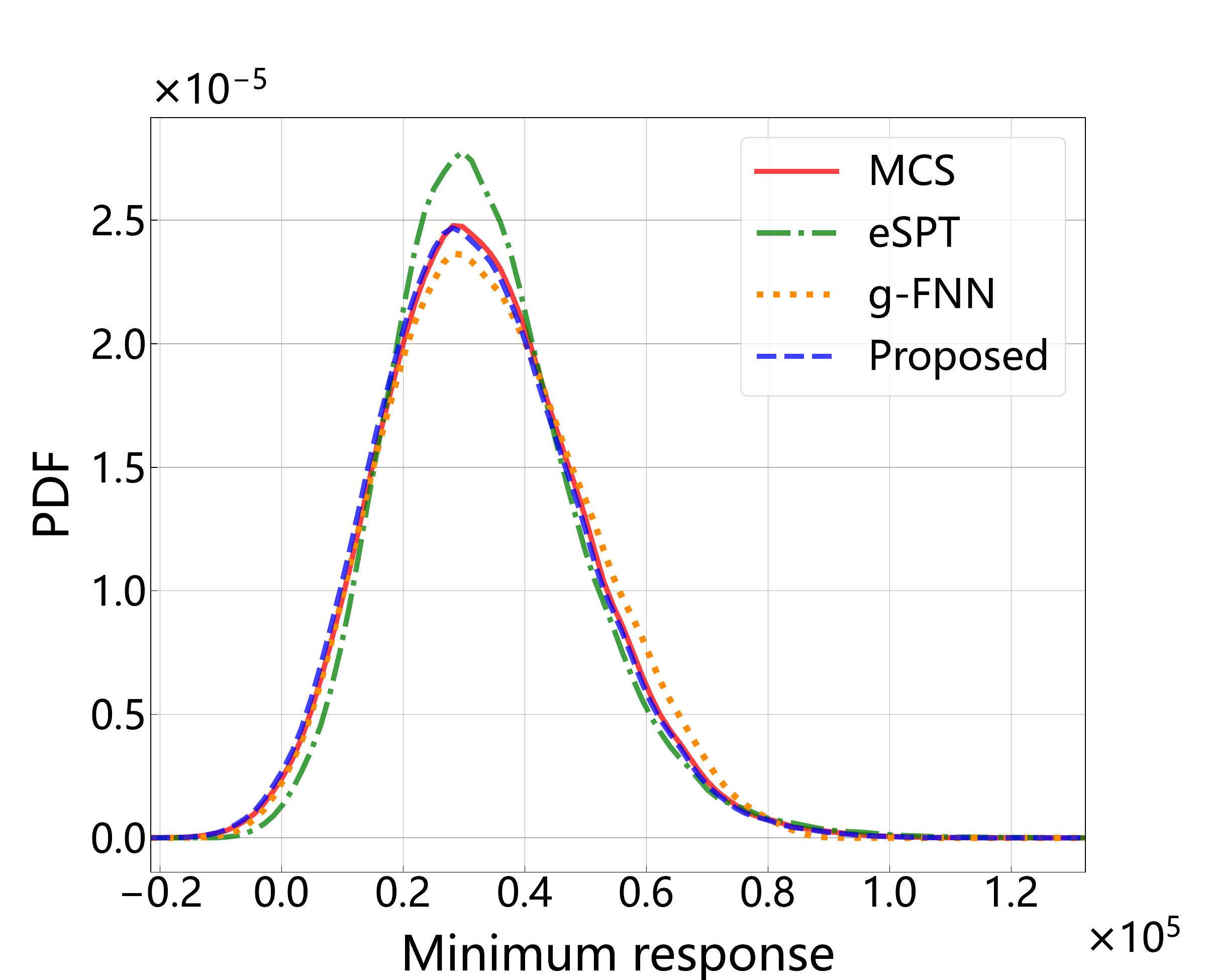}}
\caption{PDFs of minimum responses obtained from different method}
\label{fig15}
\end{figure}

In this case study, the proposed approach is further evaluated under limited-data conditions: the model is trained on only the first 50 time nodes of each 100-node sequence, yet must predict responses across the entire time horizon. As shown in Fig.~\ref{fig14}, the predicted minimum responses for all MCS samples closely match the actual values over the full duration. This demonstrates the DDF-LSTM model's capability to accurately extrapolate system behavior beyond the training time range.
Under the same conditions, different methods are tested for comparison. The results, summarized in Fig.~\ref{fig15} and Table~\ref{table8}, show that both the eSPT and g-FNN methods produce higher relative errors. In this case, since the limit state function explicitly depends on the time variable $t$, the eSPT method fails to effectively capture the degradation of system performance over time, thereby exhibiting limitations in forecasting future behavior when data is insufficient. The reference reliability computed by direct MCS is 0.98956. With only 12 time series samples used for training, the proposed method achieves a reliability estimate of 0.98815, corresponding to a relative error of 0.14249\%. Although reducing the length of training data may introduce accumulated errors and slightly reduce prediction accuracy for unknown time intervals, the proposed method still performs accurate reliability predictions over future time intervals.

\begin{table}
\caption{Comparison of reliability estimates under limited training data}
\begin{center}
\label{table8}
\resizebox{\columnwidth}{!}{%
\begin{tabular}{l c c c}
\hline
Method & \makecell{No. of Time-\\series data} & Reliability & Relative error (\%) \\
\hline
MCS & $10^6$ & 0.98956 & — \\
eSPT & $\sim$89 & 0.99647 & 0.69829 \\
g-FNN & 12 & 0.99234 & 0.28093 \\
Proposed & 12 & 0.98815 & 0.14249 \\
\hline
\end{tabular}
}
\end{center}
\end{table}

\subsection{Case Study III: Roof Truss Structure}
\begin{figure}
\centerline{\includegraphics[width=\columnwidth]{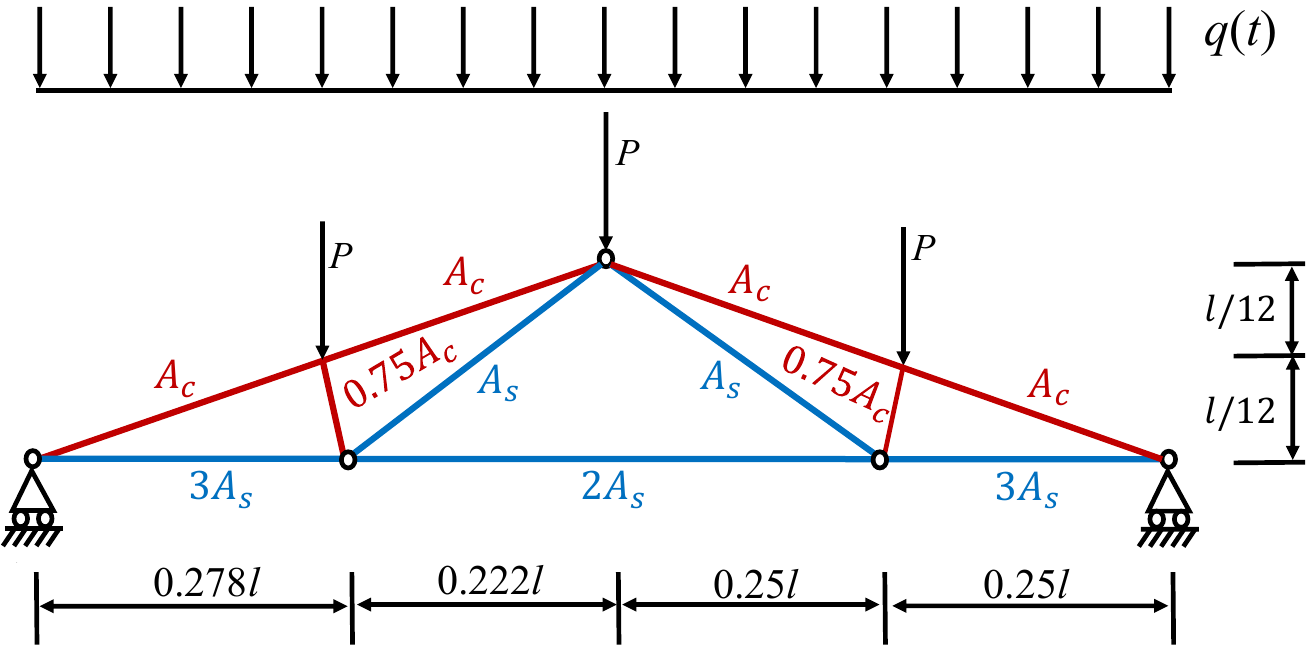}}
\caption{Schematic diagram of the roof truss structure}
\label{fig16}
\end{figure}

This third validation scenario employs a modified roof truss \cite{pan2017sliced,shi2018time} to demonstrate the methodology's efficacy. The geometry and boundary conditions of the structure are illustrated in Fig.~\ref{fig16}. The top boom and all compression members are composed of concrete, while the bottom boom and all tension members are reinforced with steel. A uniformly distributed time-dependent load $q(t)$ is applied to the structure and transformed into equivalent nodal load $P=q(t)l/4$, where $l$ denotes the length of the bottom boom. The load $q(t)$ follows a non-stationary Gaussian process defined by the following statistical characteristics:

\begin{equation}
\mu_q(t)=20000+100\sin(0.62t)
\end{equation}

\begin{equation}
\sigma_q(t)=1400+10\cos(0.62t)
\end{equation}

\begin{equation}
\rho_q(t_1,t_2)=\exp\left(-(t_1-t_2)^2\right)
\end{equation}
where $\mu_q(t)$, $\sigma_q(t)$, $\rho_q(t)$ denote the mean function, standard deviation function, and autocorrelation function, respectively. The limit state function involves five random variables and one stochastic process, expressed as

\begin{equation}
    \begin{split}
        G(\mathbf{X},\mathbf{Z}(t)) &=0.03-\Delta C \\
        &=0.03-\frac{ql^2}{2}(\frac{3.81}{A_C E_C}+\frac{1.13}{A_S E_S})
    \end{split}
\label{eq_G3}
\end{equation}
where $A_C$ and $E_C$ denote the cross-sectional area and elastic modulus of the concrete members, respectively, and $A_S$ and $E_S$ correspond to those of the steel members. $\Delta C$ represents the vertical deflection of the roof peak node. Structural failure is considered to occur when the vertical deflection exceeds 3 cm. The total analysis period is set as [0,10] years, uniformly discretized into 100 time nodes. All parameters are treated as independent as listed in Table~\ref{table9}.

\begin{table}
\caption{Statistical properties of input variables in the roof truss case study}
\label{table9}
\begin{center}
\resizebox{\columnwidth}{!}{%
\begin{tabular}{l c c c}
\hline
Input variable & Distribution & Mean & \makecell{Standard \\deviation} \\
\hline
$l\,(\text{m})$ & Normal & 12 & 0.12 \\
$A_S\,(\text{m}^2)$ & Normal & $9.82\times10^{-4}$ & $5.892\times10^{-5}$ \\
$E_S\,(\text{Pa})$ & Normal & $1\times10^{11}$ & $6\times10^9$ \\
$A_C\,(\text{m}^2)$ & Normal & $4\times10^{-2}$ & $4.8\times10^{-3}$ \\
$E_C\,(\text{Pa})$ & Normal & $2\times10^{10}$ & $1.2\times10^9$ \\
$q(t)\,(\text{N/m})$ & \makecell{Nonstationary \\Gaussian} 
& \makecell{$20000+$ \\ $100\sin(0.62t)$} & \makecell{$1400+$ \\ $10\cos(0.62t)$} \\
\hline
\end{tabular}
}%
\end{center}
\end{table}

\begin{figure}
\centerline{\includegraphics[width=\columnwidth]{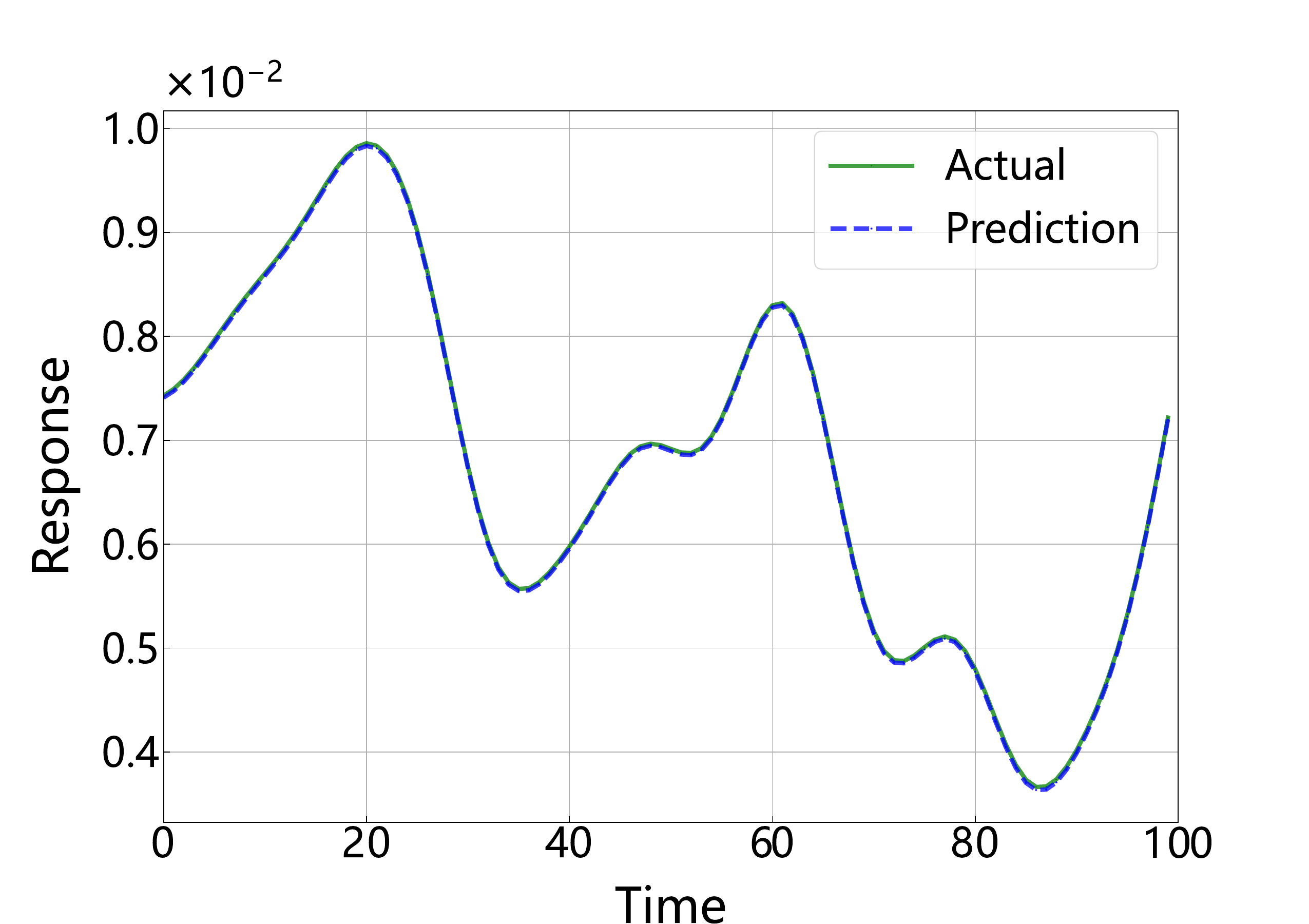}}
\caption{Comparison of DDF-LSTM predicted and actual responses for the 1st MCS sample}
\label{fig17}
\end{figure}

\begin{figure}
\centerline{\includegraphics[width=\columnwidth]{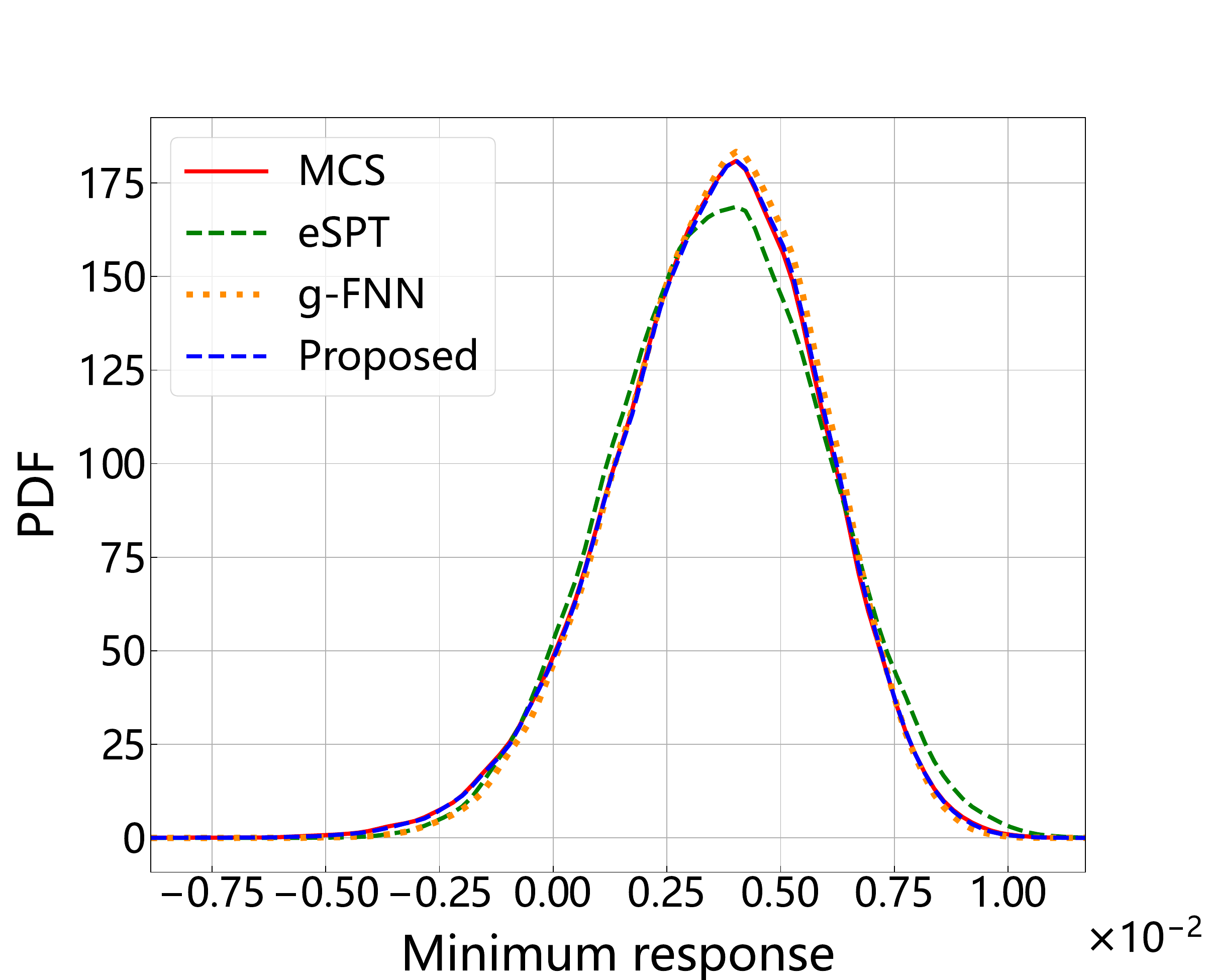}}
\caption{PDF of the minimum responses using different methods}
\label{fig18}
\end{figure}

To apply the proposed method, 20 random realizations of input variables are generated. Time-dependent responses are then computed according to Eqn.~(\ref{eq_G3}). The DDF-LSTM model is constructed with 10 neurons in the hidden layer, and $10^6$ MCS samples are produced to estimate the time-dependent reliability over the interval [0, 10] years. To demonstrate the model's ability to capture the global time-dependent response behavior, Fig.~\ref{fig17} compares the predicted and actual responses for the 1st MCS sample. As shown, the predicted curves closely align with the actual responses across the entire time interval, confirming the model's accuracy. A further comparison between different methods is presented in Table.~\ref{table10} and Fig.~\ref{fig18}. It can be observed that the PDF obtained by the proposed approach nearly overlaps with that of the direct MCS results. Notably, the proposed method achieves the lowest reliability estimation error using only 20 time series, highlighting its higher data utilization efficiency. These findings validate the method's efficacy for time-variant reliability assessment under non-stationary Gaussian process constraints. 

The training times for all methods, as listed in Table~\ref{table10}, are obtained on a testing platform equipped with an Intel i5-13400 CPU. Among them, the g-FNN model requires multiple independent training stages, iterative optimizations, and sequential hyperparameter tuning due to its multi-stage modeling framework. This non–end-to-end and complex process results in a significantly higher computational cost. In contrast, the proposed method adopts an end-to-end single-model architecture that directly learns the complex mapping from static random variables and time-varying stochastic processes to system responses, thereby avoiding the additional overhead caused by multi-stage training and model switching. Although the DDF-LSTM model inherently involves a certain degree of parameter complexity when handling sequential data, its total training time remains notably lower than that of g-FNN and even outperforms the relatively simple eSPT method. In summary, while ensuring high predictive accuracy, the proposed approach demonstrates a clear advantage in computational efficiency owing to its concise and effective modeling strategy.

\begin{table}
\caption{Comparison of reliability estimates using different prediction methods}
\begin{center}
\label{table10}
\resizebox{\columnwidth}{!}{%
\begin{tabular}{l c c c c}
\hline
Method & \makecell{No. of Time-\\series data} & Reliability & Relative error (\%) & Time (s) \\
\hline
MCS & $10^6$ & 0.93415 & — & — \\
eSPT & $\sim$99 & 0.94126 & 0.76112 & 48.2 \\
g-FNN & 20 & 0.94797 & 1.47942 & 415.22 \\
Proposed & 20 & 0.93533 & 0.12632 & 41.13 \\
\hline
\end{tabular}
}
\end{center}
\end{table}

\begin{figure}
\centerline{\includegraphics[width=\columnwidth]{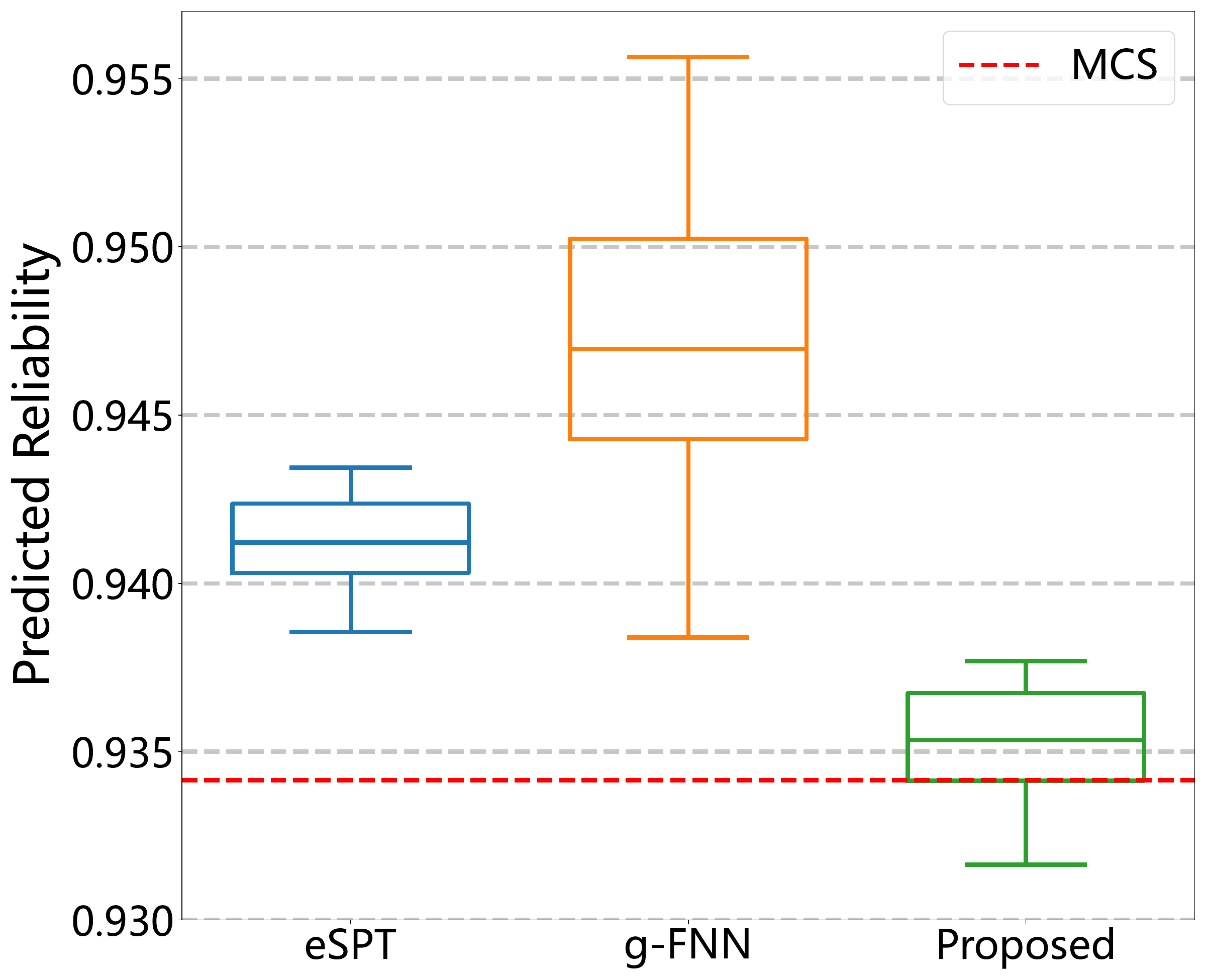}}
\caption{Statistical comparison of predicted reliability across 20 repeated runs}
\label{fig19}
\end{figure}

To validate the proposed method's robustness, 20 independent runs are conducted using randomly generated datasets for each comparison method. Fig.~\ref{fig19} summarizes the reliability estimates, with the red dashed line indicating the reference value from direct MCS. Among the compared methods, the proposed DDF-LSTM approach yields reliability estimates that are most closely aligned with the MCS benchmark and exhibit a relatively compact interquartile range, indicating both high prediction accuracy and strong consistency across trials. In contrast, g-FNN exhibits a wide spread in reliability estimates, suggesting greater sensitivity to training data variability and lower robustness. Although eSPT shows relatively consistent results, its predictions systematically overestimate the reliability compared to MCS. Overall, the results confirm that the proposed DDF-LSTM method not only achieves reliable and stable predictions with fewer training samples, but also offers enhanced generalization and robustness for time-dependent reliability analysis.

\subsection{Case Study IV: A Stone Arch Bridge Under Hurricane Load}
Hurricanes are among the most damaging extreme environmental hazards, frequently causing widespread disruption, severe economic losses, and safety risks to infrastructure systems. As critical components of transportation networks, bridges are particularly susceptible to extreme wind effects. Accurately understanding their structural responses under hurricane-induced loading is therefore crucial for improving risk mitigation, ensuring operational safety, and enhancing network resilience.

\begin{figure}
\centerline{\includegraphics[width=\columnwidth]{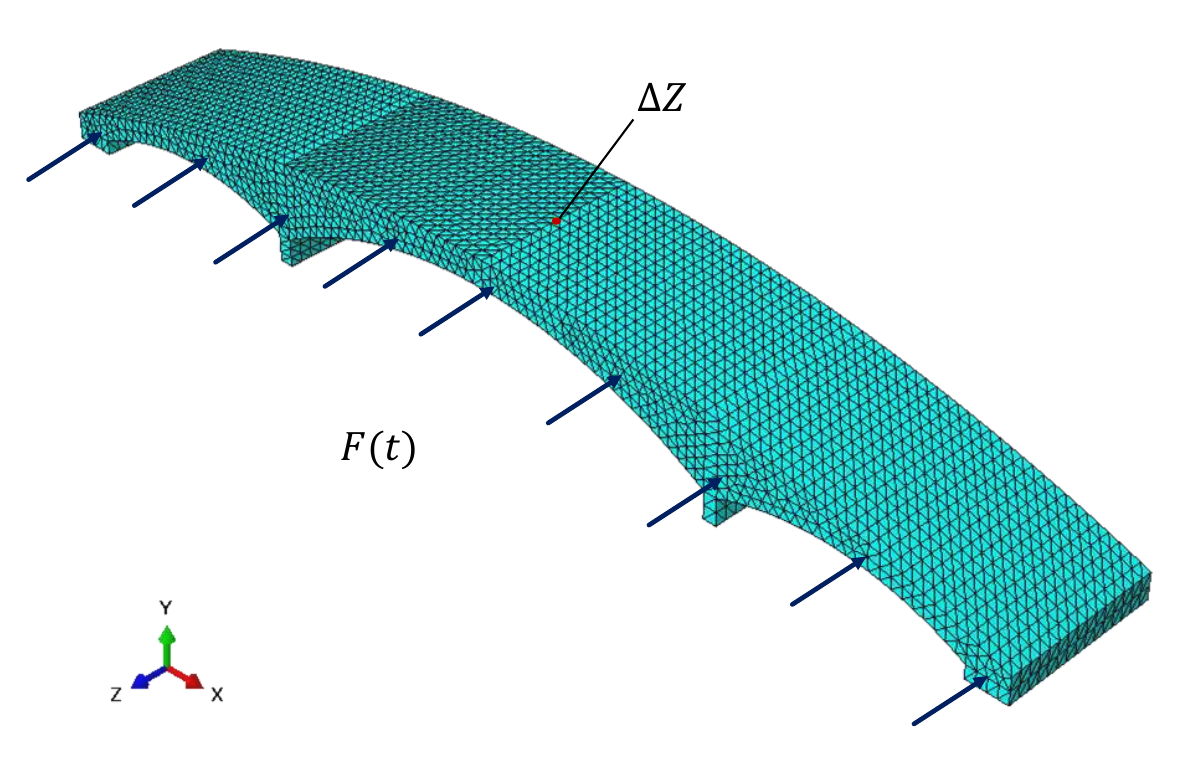}}
\caption{Finite element model of the stone arch bridge}
\label{fig20}
\end{figure}

\begin{figure*}
\centerline{\includegraphics[width=5in]{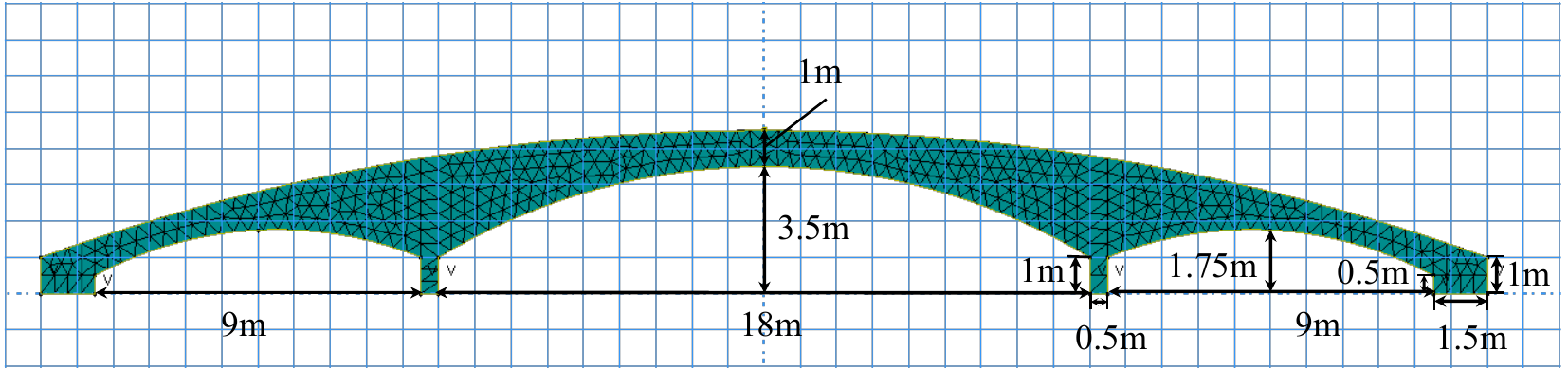}}
\caption{Structural geometry and dimensions of the stone arch bridge}
\label{fig21}
\end{figure*}

\begin{figure*}
\centerline{\includegraphics[width=5in]{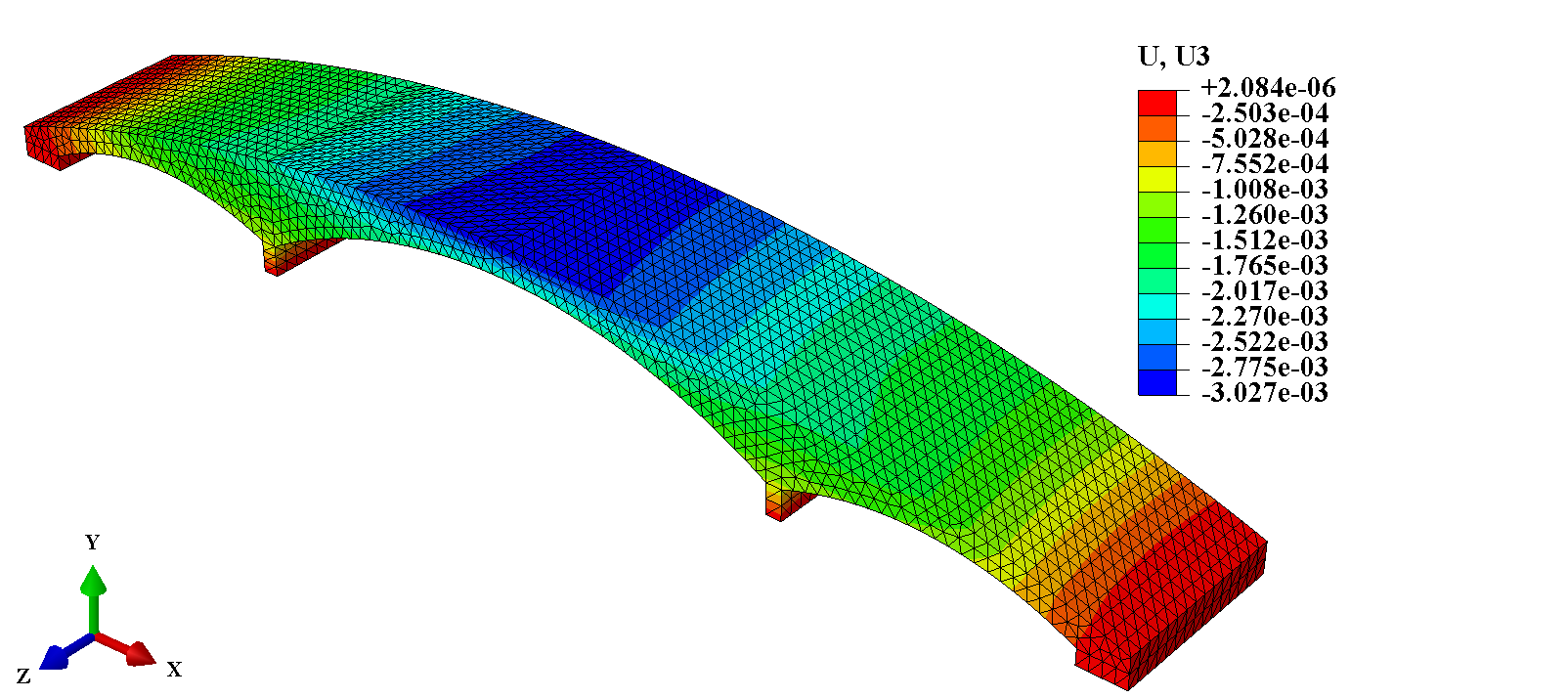}}
\caption{Deformation of the stone arch bridge at $t=5\,\text{s}$}
\label{fig23}
\end{figure*}

In this case study, a stone arch bridge subjected to hurricane loading is modified based on reference \cite{shi2019adaptive} to analyze. A detailed finite element model of the structure is developed using ABAQUS, as shown in Fig.~\ref{fig20}. The bridge is symmetric with a total width of $L=6\,\text{m}$, and its geometric profile is illustrated in Fig.~\ref{fig21}. The bottom surface of the structure is fully fixed, and a time-dependent periodic hurricane load $F(t)$ acts uniformly on the bridge surface in the negative Z-direction. The loading function $F(t)$ is defined as:
\begin{equation}
F(t)=F_0+0.4F_0\cos(2\pi t)+0.6F_0\cos(4\pi t)
\end{equation}
where $F_0$ is a normally distributed variable and $t \in [0,5]\text{s}$ is uniformly discretized into 100 nodes for analysis. The material density ($\rho$), elastic modulus ($E$), Poisson's ratio ($\upsilon$), and load ($F_0$) are considered as input variables in the analysis. Table~\ref{table11} presents their corresponding statistical properties. The absolute displacement $|\Delta Z|$ at the middle node of the bridge deck, along the direction $Z$ axis, is selected as the system response for reliability evaluation. A failure event is considered to occur if $|\Delta Z|$ exceeds 0.032m. Accordingly, the limit state function can be formulated as:
\begin{equation}
G(\mathbf{X},\mathbf{Z}(t),t)=0.032-|\Delta Z|
\label{eq_G4}
\end{equation}
where $\mathbf{X}$ represents the set of time-independent input random variables. When all input parameters are assigned their mean values, the bridge deformation at $t=5\,\text{s}$ is depicted in Fig.~\ref{fig23}.

\begin{table}
\caption{The distribution parameters of the inputs}
\begin{center}
\label{table11}
\resizebox{\columnwidth}{!}{%
\begin{tabular}{l l l l}
\hline
Input variable & Distribution & Mean & Standard deviation \\
\hline
$\rho\,(\text{Kg/m}^3)$ & Normal & $2.43\times10^3$ & $1.215\times10^2$ \\
$E\,(\text{Pa})$ & Normal & $5.23\times10^7$ & $2.615\times10^6$ \\
$\upsilon$ & Normal & 0.3 & 0.015 \\
$F_0\,(\text{Pa})$ & Normal & $1.5\times10^3$ & 75 \\
\hline
\end{tabular}
}
\end{center}
\end{table}

\begin{figure*}
\centering  
    \begin{subfigure}[b]{\columnwidth}
        \centering
        \includegraphics[width=\linewidth]{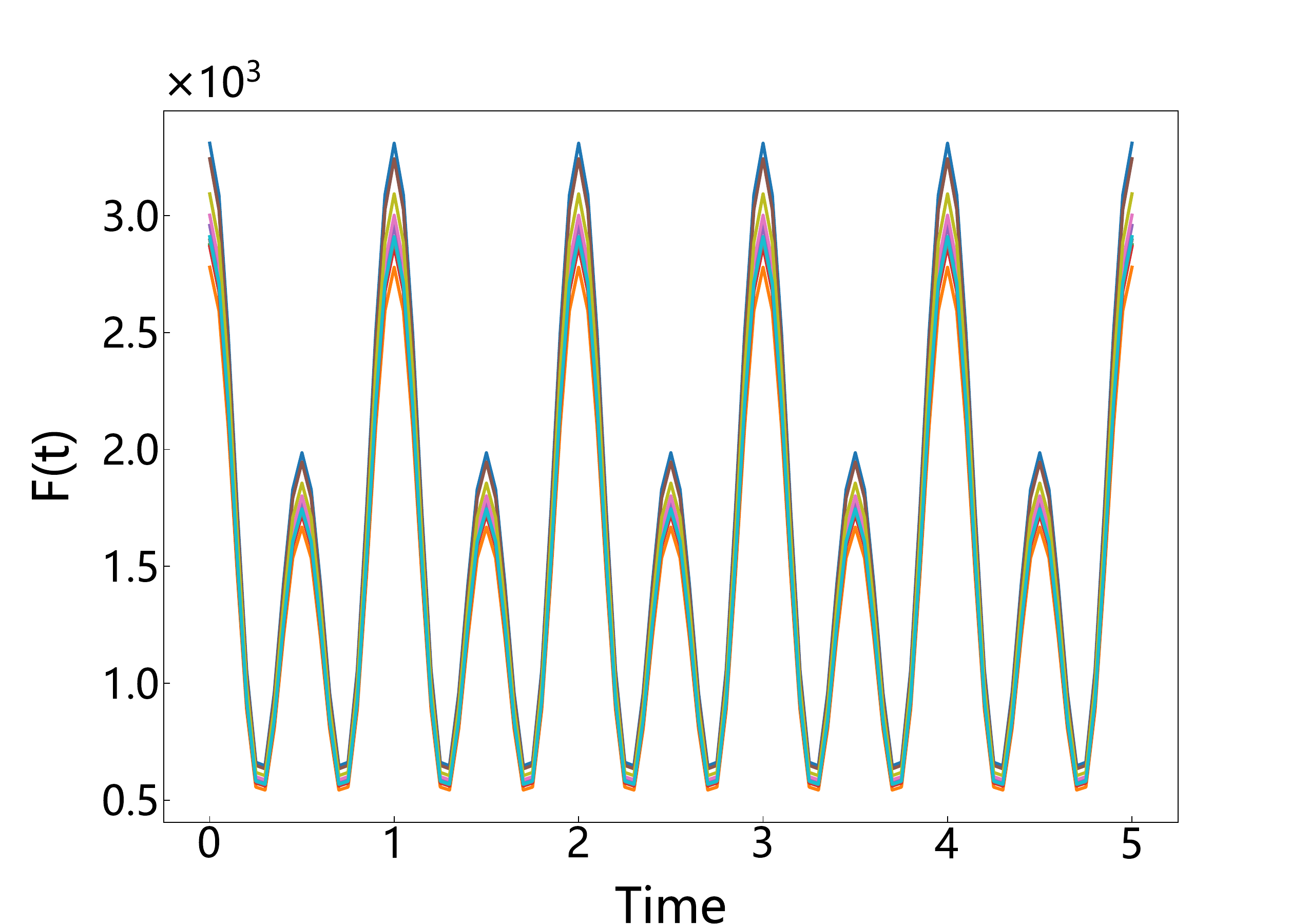}
        \caption{}
    \end{subfigure}
    \begin{subfigure}[b]{\columnwidth}
        \centering
        \includegraphics[width=\linewidth]{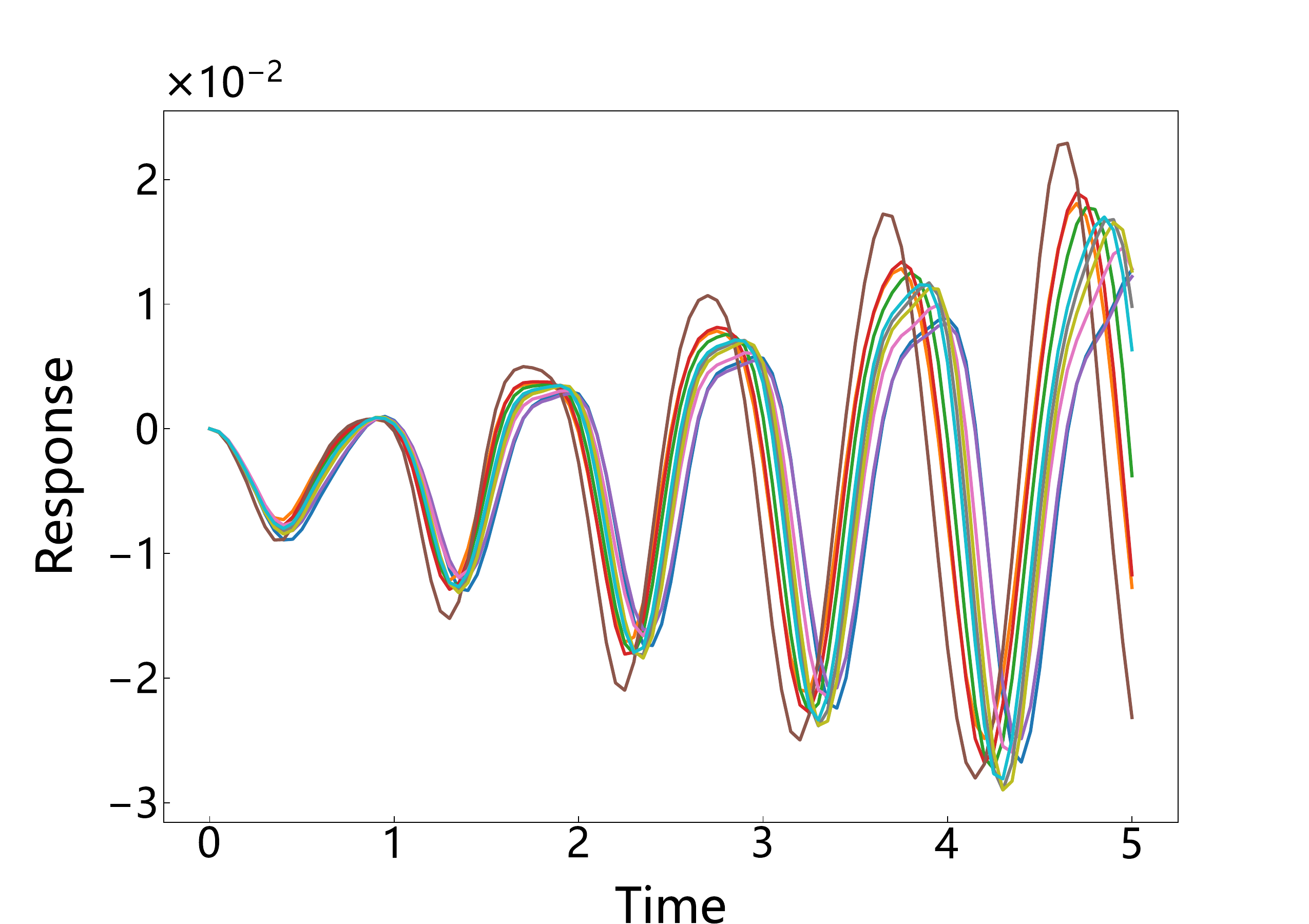}
        \caption{}
    \end{subfigure}
\caption{Ten sample realizations of stochastic processes and responses}
\label{fig24}
\end{figure*}

To implement the proposed approach, 30 training samples are generated, each consisting of four random variables and one stochastic process. By applying the finite element model, the corresponding displacement $\Delta Z$ can be obtained, producing the time-dependent response data according to Eqn.~(\ref{eq_G4}). Ten sample realizations of $F(t)$ are plotted in Fig.~\ref{fig24}(a), with the associated time-dependent responses shown in Fig.~\ref{fig24}(b). It can be observed that as the hurricane load fluctuates periodically, the displacement at the node exhibits increasingly larger oscillations over time due to the dynamic behavior.

\begin{figure}
\centerline{\includegraphics[width=\columnwidth]{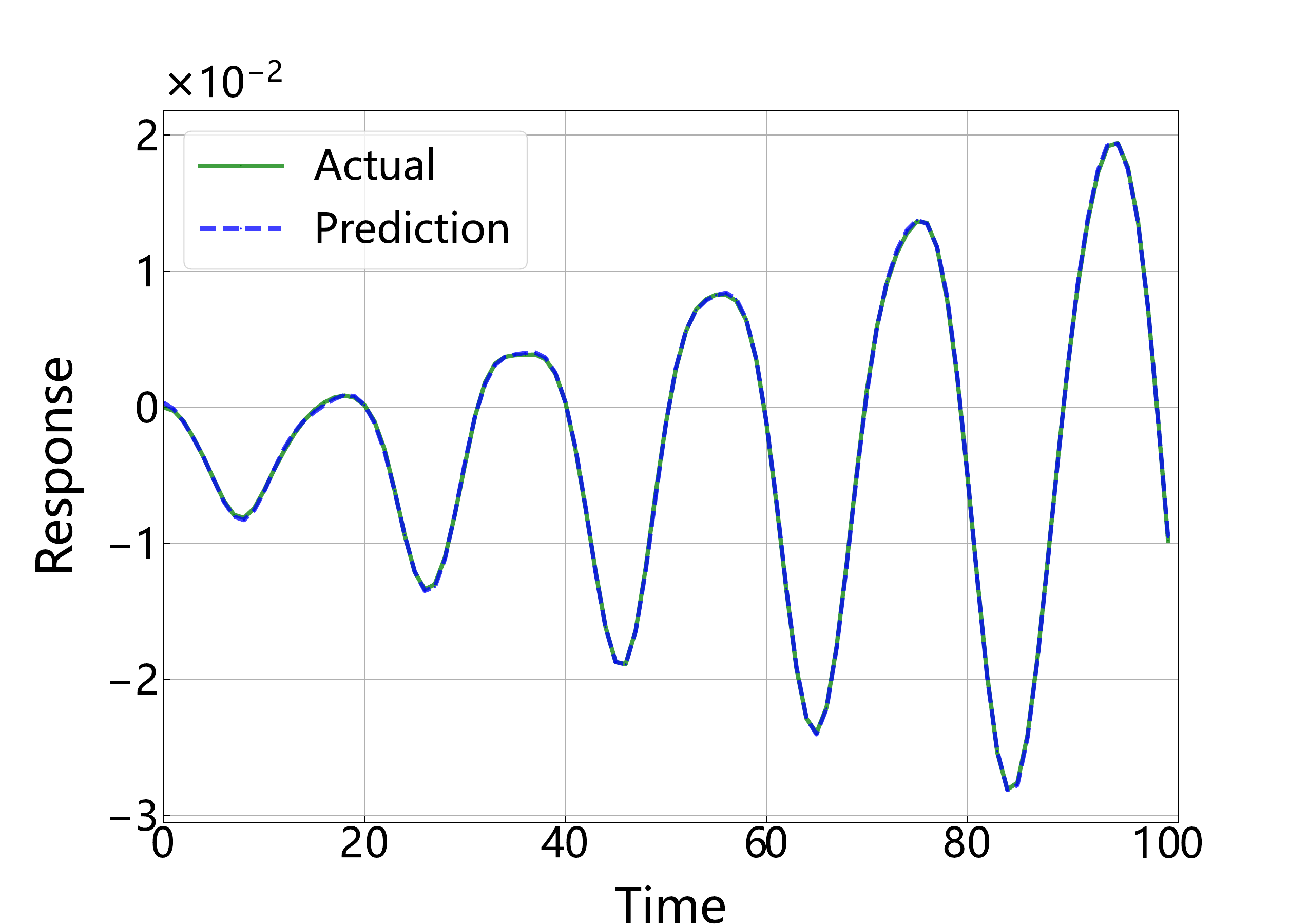}}
\caption{Comparisons between actual and predicted time-dependent responses of the 125th MCS sample}
\label{fig25}
\end{figure}

In this case, the finite element computations are computationally expensive. Therefore, only $10^4$ MCS samples are generated for demonstration purposes, considering practical computational constraints. The predicted displacement responses for the 125th MCS sample are compared with the corresponding reference responses in Fig.~\ref{fig25}. The predicted time-series data closely match the actual values throughout the entire time interval, confirming the accuracy of the DDF-LSTM model.

\begin{figure}
\centerline{\includegraphics[width=\columnwidth]{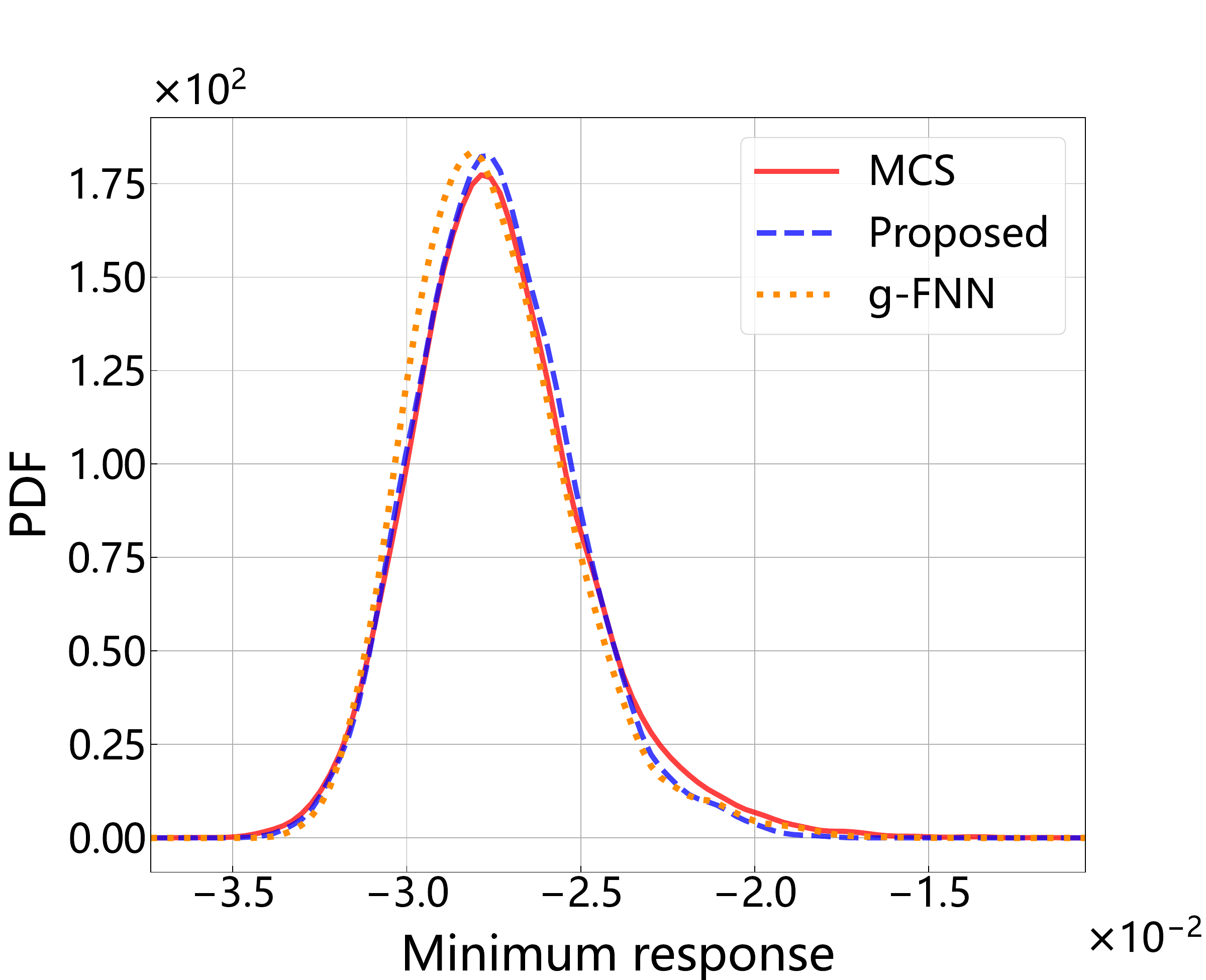}}
\caption{PDF curves of minimum structural responses obtained by different methods}
\label{fig26}
\end{figure}

To optimize failure surface detection, eSPT using the Maximum Confidence Enhancement (MCE) strategy for iteratively updating the Kriging model. However, calculating Expected Improvement (EI) values requires generating discrete MCS samples and obtaining the corresponding performance function responses, which can be computationally expensive due to the large number of finite element analyses involved. Additionally, the g-FNN method is adopted for comparison. Fig.~\ref{fig26} presents the PDFs of the minimum responses for all $10^4$ MCS samples as estimated by different methods. Table~\ref{table12} summarizes the reliability estimates. The proposed DDF-LSTM approach achieves a reliability estimate of 0.9865, with a relative error of 0.21333\%, demonstrating its superior capability in capturing extreme response characteristics and maintaining high accuracy under limited data conditions. Although g-FNN can approximate the overall distribution, it exhibits deviations in the peak position and distribution width of the PDF, leading to increased errors in reliability prediction. These findings demonstrate that the proposed approach achieves excellent precision in evaluating the time-dependent reliability of complex systems under uncertainty.

\begin{table}[t]
\caption{Comparison of time-dependent reliability estimates}
\begin{center}
\label{table12}
\resizebox{\columnwidth}{!}{%
\begin{tabular}{l c c c}
\hline
Method & \makecell{No. of Time-\\series data} & Reliability & Relative error (\%) \\
\hline
MCS & $10^4$ & 0.9844 & — \\
g-FNN & 30 & 0.9916 & 0.73141 \\
Proposed & 30 & 0.9865 & 0.21333 \\
\hline
\end{tabular}
}
\end{center}
\end{table}

\section{Conclusion}
\label{sect_conclusion}

This paper proposes a dualdomain fused LSTM model for efficient and accurate timedependent reliability analysis. The proposed framework seamlessly integrates the modeling of time-independent random variables and time-dependent stochastic processes within a unified neural architecture. By embedding static variables into the initial LSTM states and fusing temporal hidden states with original static features via a fully connected layer, this approach effectively mitigates the inefficiency and error propagation often associated with multi-model strategies. Furthermore, a custom loss function is introduced to emphasize accuracy in predicting minimum responses through focusing on the extreme value. The effectiveness and generality of the DDF-LSTM model are demonstrated through four diverse case studies. The results consistently indicate that the proposed method achieves high predictive accuracy with minimal training data, significantly outperforming conventional surrogate models. Moreover, the model exhibits strong generalization capabilities for unseen time intervals and ensures robust reliability estimates with minimal variance. In summary, the DDF-LSTM framework offers a computationally efficient, accurate, and robust solution for time-dependent reliability assessment, particularly for problems characterized by high-dimensional stochasticity and limited data availability.

\bibliographystyle{asmems4}
\bibliography{ref}

\end{document}